\documentclass[letterpaper, 10 pt, conference]{ieeeconf}  

\IEEEoverridecommandlockouts                              

\usepackage[utf8]{inputenc} 
\usepackage[T1]{fontenc}    
\usepackage{hyperref}       
\usepackage{url}            
\usepackage{booktabs}       
\usepackage{amsfonts}       
\usepackage{nicefrac}       
\usepackage{microtype}      
\usepackage{algorithm}
\usepackage[noend]{algpseudocode}
\usepackage{graphicx}
\usepackage{subcaption}
\usepackage{tikz}
 \usepackage{amssymb}
\usepackage{booktabs} 
\newtheorem{theorem}{Theorem}[section]
\newtheorem{lemma}[theorem]{Lemma}

\newtheorem{definition}[theorem]{Definition}
\newtheorem{assumption}{Assumption}

\title{\LARGE \bf
Adaptive Policy Transfer in Reinforcement Learning
}

\author{Girish Joshi and Girish Chowdhary
\thanks{*This work was supported in part by ONR MURI Navy N00014-19-1-2373, and AFOSR YIP FA9550-15-1-0146}
\thanks{Coordinated Sciences Laboratory, University of Illinois, Urbana-Champaign.
        {\tt\small girishj2@illinois.edu, girishc@illinois.edu}}%
}

\begin{document}

\maketitle
\thispagestyle{empty}
\pagestyle{empty}

\begin{abstract}
Efficient and robust policy transfer remains a key challenge for reinforcement learning to become viable for real-wold robotics. Policy transfer through warm initialization, imitation, or interacting over a large set of agents with randomized instances, have been commonly applied to solve a variety of Reinforcement Learning tasks. However, this seems far from how skill transfer happens in the biological world: Humans and animals are able to quickly adapt the learned behaviors between similar tasks and learn new skills when presented with new situations. Here we seek to answer the question: Will learning to combine adaptation and exploration lead to a more efficient transfer of policies between domains? We introduce a principled mechanism that can \textbf{``Adapt-to-Learn"}, that is adapt the source policy to learn to solve a target task with significant transition differences and uncertainties.  We show 
that the presented method learns to seamlessly combine learning from adaptation and exploration and leads to a robust policy transfer algorithm with significantly reduced sample complexity in transferring skills between related tasks.

\end{abstract}
\section{Introduction}
\label{introduction}
Lack of principled mechanisms to quickly and efficiently transfer learned policies when the robot dynamics or tasks significantly change, has become the major bottleneck in Reinforcement Learning (RL). This inability to transfer policies is one major reason why RL has still not proliferated physical application like robotics \cite{peng2016terrain,peng2017deeploco,liu2017learning,heess2017emergence}. This lack of methods to efficiently transfer robust policies forces the robot to learn to control in isolation and from scratch, even when only a few parameters of the dynamics change across tasks (e.g. walking with different weight or surface friction, etc.). This is both computation and sample expensive, and also makes simulation-to-real transfer difficult. 

Adapt-to-Learn (ATL) is inspired by the fact that  adaptation of behaviors  combined with learning through experience is a primary mechanism of learning in biological creatures \cite{krakauer2011human,fryling2011understanding}. Imitation Learning (IL) \cite{duan2017one, zhu2018dexterous} also seems to play a very crucial part in skill transfer, and as such has been widely studied in RL. In control theory, adaptation 
has been 
typically studied for tracking problems on deterministic dynamical systems with well-defined reference trajectories 
\cite{aastrom2013adaptive,chowdhary2013rapid}. Inspired by learning in biological world, we seek to answer the question: Will combining adaptation of transferred policies and learning from exploration lead to more efficient policy trasnfer?
In this paper, we show that the ability to adapt and incorporate further learning can be achieved through optimizing over combined environmental and intrinsic adaptation rewards. Learning over the combined reward feedback, we ensure that the agent quickly adapts and also learns to acquire skills beyond what the source policy can teach. Our empirical results show that the presented policy transfer algorithm is able to adapt to significant differences in the transition models, which otherwise, using Imitation Learning or Guided Policy Search (GPS)~\cite{levine2013guided} would fail to produce stable solution. We posit that the presented method can be the foundation of a  broader class of RL algorithms that can choose seamlessly between learning through exploration and supervised adaption using behavioral imitation.

\section{Related work}  
Deep Reinforcement Learning (D-RL) has recently enabled agents to learn policies for complex robotic tasks in simulation \cite{peng2016terrain,peng2017deeploco,liu2017learning,heess2017emergence}. However, D-RL applications to robotics have been plagued by the curse of sample complexity.  
Transfer Learning (TL) seeks to mitigate this learning inefficiency of D-RL \cite{taylor2009transfer}.
A significant body of literature on transfer in RL is focused on initialized RL in the target domain using learned source policy; known as jump-start/warm-start methods \cite{taylor2005value,ammar2012reinforcement,ammar2015unsupervised}. Some examples of these transfer architectures include transfer between similar tasks \cite{banerjee2007general}, transfer from human demonstrations \cite{peters2006policy} and transfer from simulation to real \cite{peng2017sim,ross2011reduction,yan2017sim}. However, warm-start based approaches do not always work in transfer for RL, even when similar warm-starts are quite effective in supervised learning.
Efforts have also been made in exploring accelerated learning directly on real robots, through Guided Policy Search (GPS) \cite{levine2015learning} and parallelizing the training across multiple agents using meta-learning \cite{levine2016learning,nagabandi2018neural,zhu2018dexterous}, but these approaches are prohibitively sample expensive for real-world robotics. Sim-to-Real transfers have been widely adopted in the recent works and can be viewed as a subset of same domain transfer problems. Daftry et al. \cite{daftry2016learning} demonstrated the policy transfer for control of aerial vehicles across different vehicle models and environments. Policy transfer from simulation to real using an inverse dynamics model estimated interacting with the real robot is presented by \cite{christiano2016transfer}.
The agents trained to achieve robust policies across various environments through learning over an adversarial loss is presented in  \cite{wulfmeier2017mutual}. Here we present a general architecture capable of transferring policies across MDPs with significantly different transition dynamics. 
\section{Preliminaries and Problem Specification}
Consider an infinite horizon MDP defined as a tuple $\mathcal{M} = (\mathcal{S},\mathcal{A},\mathcal{P},\mathcal{R},\rho_0,\gamma)$, where $\mathcal{S}$ denote set of continuous states; 
$\mathcal{A}$ is a set of continuous bounded actions, $\mathcal{P}: \mathcal{S} \times \mathcal{A} \times \mathcal{S}\to \mathbb{R}_+$ is the state transition probability distribution of reaching $s'$ upon taking action $a$ in $s$, $\rho_0: \mathcal{S} \to \mathbb{R}_+$ is the distribution over initial states $s_0$ and $\mathcal{R}:\mathcal{S} \times \mathcal{A} \to \mathbb{R}_+$ is deterministic reward function, s.t. $\mathcal{R} \in [0,1]$ and $\gamma$ is the discount factor s.t $0 < \gamma \leq 1$.

Let $\pi(a|s): \mathcal{S} \times \mathcal{A} \to  [0,1]$ be stochastic policy over continuous state and action space. The action from a policy is a draw from the distribution $a_i \sim \pi(.|s_i)$. The agent's goal is to find a policy $\pi^*$ which maximize the total return. The total return under a policy $\pi$ is given as,
\begin{equation}
    \eta_{\pi}(s_0) = \mathbb{E}_{s_0,a_0, \dots}\left(\sum_{t=0}^\infty \gamma^t r(s_t,a_t)\right).
\end{equation}
where, $s_0 \sim \rho_0$, $a_t \sim \pi(.|s_t)$ and $s_{t+1} \sim p(.|s_t, a_t)$.

We will use the following definition of the state value function $V^{\pi}$ and state-action value function $Q^{\pi}$ defined under any policy $\pi$
$$
V^{\pi}(s_t) =  \mathbb{E}_{a_t, s_{t+1},a_{t+1}, \dots}\left(\sum_{i=0}^\infty \gamma^i r(s_{i+t},a_{i+t})\right),
$$
$$
Q^{\pi}(s_t, a_t) =  \mathbb{E}_{s_{t+1},a_{t+1}, \dots}\left(\sum_{i=0}^\infty \gamma^i r(s_{i+t},a_{i+t})\right).
$$

With the quick overview of the preliminaries, we now specify the problem of policy transfer studied here: Consider a source MDP $\mathcal{M}^S = (\mathcal{S},\mathcal{A},\mathcal{P},\mathcal{R},\rho_0,\gamma)^S$, and a target MDP  $\mathcal{M}^T = (\mathcal{S},\mathcal{A},\mathcal{P},\mathcal{R},\rho_0,\gamma)^T$, each with its own state-action space and transition model respectively. 
We will mainly focus on the problem of same domain transfer in this paper, where the state and action space are analogous $\mathcal{S}^{(S)} = \mathcal{S}^{(T)} = \mathcal{S} \subseteq \mathbb{R}^m$ and $\mathcal{A}^{(T)} = \mathcal{A}^{(S)} = \mathcal{A} \subseteq \mathbb{R}^k$, but 
the source and target state transition models differ significantly 
due to 
unmodeled dynamics or external environment interactions. We are concerned with this problem because such policy adaptation in the presence of significant transition model change happens quite often in robotics, e.g. change in surface friction, payload carried by robot, modeling differences, or component deterioration. 
Note that extension to cross-domain transfer could be achieved with 
domain alignment techniques such as manifold alignment (MA) \cite{wang2009manifold}, see \cite{joshi2018cross} for a model-based policy transfer method for RL with MA.

Let $\pi^{*}$ be a parameterized optimal stochastic policy for source MDP $\mathcal{M}^S$. The source policy $\pi^{*}$ can be obtained using any available RL methods \cite{sutton2000policy, schulman2015trust, schulman2017proximal}.
We use Proximal Policy Optimization (PPO) \cite{schulman2017proximal} algorithm to learn a optimal source policy on a unperturbed source task. A warm-started PPO-TL \cite{ammar2015unsupervised} policy, trained over a perturbed target task and ideal imitation learning is used as the TL solutions against which our proposed ATL policy is compared. 

\section{Adaptive Policy Transfer}
\label{Section:TATL}
In this paper, we approach the problem of transfer learning for RL through adaptation of previously learned policies. Our goal in the rest of this paper is to show that an algorithm that can judiciously combine adaptation and learning is capable of avoiding brute force random exploration to a large extent and therefore can be significantly less sample expensive.

Towards this goal, our approach is to enable the agent to learn the optimal mixture of adaptation (which we term as \textit{behavioral imitation}) and learning from exploration. Our method differs from existing transfer methods that rely on warm start for TL or policy imitation \cite{zhu2018dexterous, duan2017one} in a key way: Unlike imitation learning, we do not aim to emulate the source optimal policy. Instead, we try to adapt and encourage directional search for optimal reward by mimicking the source transitions under source optimal policy as projected onto the target state space. The details of Adaptive policy Transfer is presented further in this section.

\subsection{Adapt-to-Learn: Policy Transfer in RL}

ATL has two components to policy transfer, Adaptation and Learning. 
We begin by mathematically describing our approach of adaptation for policy transfer in RL and state all the necessary assumptions in Section \ref{sec:adaptation}. We then develop the Adapt-to-Learn algorithm in Section \ref{sec:ATL} by adding the learning component by random exploration through reward mixing.

\subsubsection{Adaptation  for policy learning}\label{sec:adaptation}
Adaptation of policies aims to generate the policy $\pi_\theta(.|s)$ such that at every given $s \in \mathcal{S}^T$ the Target transition approximately mimics the Source transitions under Source optimal policy as projected onto the Target state space. We can loosely state the adaptation objective as
$
\mathcal{P}^T(.|s, \pi_\theta(a|s)) \approx \mathcal{P}^S(.|\hat{s}, \pi^*(a'|\hat{s})), \hspace{2mm}\forall s \in \mathcal{S}^T,
$.
Where $\hat{s} = \chi_T^S(s)$, i.e. target state projected onto source manifold. The mapping $\chi_T^S$ is a bijective Manifold Alignment (MA) between source and target, state-action spaces $\{\mathcal{S}, \mathcal{A}\}^S$, $\{\mathcal{S}, \mathcal{A}\}^T$. The MA mapping is required for cross-domain transfers \cite{ammar2015unsupervised,joshi2018cross}. For ease of exposition, in this paper, we focus on same domain transfer and assume MA mapping to be identity, such that $\chi_T^S = \chi_S^T = I$.

Note that our goal is not to directly imitate the policy itself, but rather use the source behavior as a baseline for finding optimal transitions in the target domain. This behavioral imitation objective can be achieved by minimizing the average KL divergence between point-wise local target trajectory deviation likelihoods $p^T(s'_{t+1}|s_t,\pi_\theta(.|s_t))$, and the source transition under source optimal policy $p^S(s'_{t+1}|s_t,\pi^*(.|s_t))$.

We define the target transition deviation likelihood as conditional probability of landing in the state $s'_{t+1}$ starting from state $s_t$  under some action $a_t \sim \pi_\theta(.|s_t)$. Unlike state transition probability, the state $s'_{t+1}$ is not the next transitioned state but a random state at which this probability is evaluated (Hence $s'_{t+1}$ need not have the time stamp).

The adaptation objective can be formalized as minimizing the average KL-divergence \cite{schulman2015trust} between source and target transition trajectories as follows,
\begin{eqnarray}
\eta_{_{KL}}(\pi_\theta, \pi^*) &=& D_{KL}(p_{\pi_\theta}(\tau)\|q_{\pi^*}(\tau)), \nonumber \\
    \eta_{_{KL}}(\pi_\theta, \pi^*) &=& \int_{\mathcal{S},\mathcal{A}} p_{\pi_\theta}(\tau)\log\left(\frac{p_{\pi_\theta}(\tau)}{q_{\pi^*}(\tau)}\right)d\tau
    \label{eq:KL_definition}
\end{eqnarray}
where $\tau = (s_0, s_1, s_2,\dots)$ is the trajectory in the target domain under the policy $\pi_{\theta}(.|s)$ defined as collection states visited starting from state $s_0 \sim \rho_0$ and making transitions under target transition model $p^T(.|s_t,a_t)$. We now define the probability of the trajectories $p_{\pi_\theta}(\tau)$ and $q_{\pi^*}(\tau)$:

The term $p_{\pi_\theta}(\tau)$ is defined as total likelihood of one-step deviations to reference states $\{s'_{t+1}\}_{t=1}^\infty$ over the trajectory $\tau$. The reference states are obtained by re-initializing the source simulator at every time ``$t$'' to the states along the trajectory $\tau$ and making optimal transitions to $\{s'_{t+1}\}_{t=1}^\infty$ using source optimal policy. The total trajectory deviation likelihood can be expressed as follows,
\begin{equation}
   p_{\pi_\theta}(\tau) = \rho(s_0)\prod_{t=0}^\infty\pi_\theta(a_t|s_t)p^T(s'_{t+1}|s_t,\pi_\theta(s_t)).
   \label{eq:sourcetraj_prob}
\end{equation}
\begin{assumption}
\label{assumption-source_simualtor}
 Access to source simulator with restart capability is assumed. That is,  source transition model can be initialized to any desired states $s_t$ and under optimal actions $a'_t \sim \pi^*(.|s_t)$ the source simulator provides next transitioned state $s'_{t+1}$.
\end{assumption}
Using Assumption-\ref{assumption-source_simualtor}, $q_{\pi^*}(\tau)$ can be defined as total probability of piece-wise transitions at every state $s_t \in \tau$ under the source optimal policy $a'_t \sim \pi^*(.|s_t)$, and the source transitions $p^S(.|s_t, a'_t)$.
\begin{equation}
    q_{\pi^*}(\tau) = \rho(s_0)\prod_{t=0}^\infty\pi^*(a'_t|s_t)p^S(s'_{t+1}|s_t,\pi^*(s_t)).
    \label{eq:targettraj_prob}
\end{equation}

Unlike the conventional treatment of KL-term in the RL literature,
the transition probabilities in the KL divergence in (\ref{eq:KL_definition}) are treated as transition likelihoods and the transitioned state $s_{t+1}$ as random variables.
The policy $\pi_\theta$ learns to mimic the optimal transitions by minimizing the KL distance between the transition likelihoods of the target agent reaching the reference states $\{s'_{i}\}_{i=1}^{\infty}$ and the source transitions evalauted at $\{s_{i}\}_{i=0}^{\infty}$. 

Using the definition of the probabilities of the trajectories under $\pi_\theta$ and $\pi^*$ ( Equation-(\ref{eq:sourcetraj_prob}) \& (\ref{eq:targettraj_prob}) ) the KL divergence (\ref{eq:KL_definition}) over the trajectory $\tau$ can be expressed as follows
\begin{eqnarray}
&&\int_{\tau} p_{\pi_\theta}(\tau)\log\left(\frac{p_{\pi_\theta}(\tau)}{q_{\pi^*}(\tau)}\right)d\tau = \\
&&
\displaystyle \mathop{\mathbb{E}}_{s_t \sim \tau}\left(\log\left(\frac{\rho(s_0)\pi(a_0|s_0)p^T(s'_1|s_0,\pi_\theta(s_0))\ldots}{\rho(s_0)\pi^*(a'_0|s_0)p^S(s'_1|s_0,\pi^*(s_0))\dots}\right)\right) \nonumber
\label{eq:dkl_logterm}
\end{eqnarray}
\begin{assumption}
\label{assumption-source_policy}
We assume that optimal source policy is available, and the probability of taking optimal actions $\pi^*(a'_t|s_t) \approx 1$, $\forall s_t \in \mathcal{S}^T$. 
\end{assumption}
We make this simplifying Assumption-\ref{assumption-source_policy} i.e. $\pi^*(a'_t|s_t) = 1 \forall s_t$ to derive a surrogate objective $\bar{\eta}_{_{KL}}(\pi_\theta, \pi^*)$ which lower bounds the true KL loss (\ref{eq:dkl_logterm}), such that, $\bar{\eta}_{_{KL}}(\pi_\theta, \pi^*) \leq \eta_{_{KL}}(\pi_\theta, \pi^*)$.

The expression for surrogate objective is defined follows,
\begin{eqnarray}
\bar{\eta}_{_{KL}}(\pi_\theta, \pi^*) &=& \displaystyle \mathop{\mathbb{E}}_{s_t \sim \tau}\left(\sum_{t=0}^\infty \zeta_t \right),
\label{eq:adaptive_learining_objective}
\end{eqnarray}
where $\zeta_t = \log\left(\frac{\pi_{\theta}(a_t|s_t)p^T(s'_{t+1}|s_t,\pi_\theta(s_t))}{p^S(s'_{t+1}|s_t,\pi^*(s_t))}\right)$ defined as the intrinsic/adaptation reward term.

We need this assumption only for deriving the expression for intrinsic reward. Assumption-\ref{assumption-source_policy} allows us to extend the proposed transfer architecture to deterministic source policy or source policy derived from a human expert; since a human teacher cannot specify the probability of choosing an action. However, in the empirical evaluation of the algorithm, the source policy used as a stochastic optimal policy. The stochastic nature of such a policy adds natural exploration, and also demonstrates that the presented method is not restricted in its application by the above assumption.

\subsubsection{Combined Adaptation and Learning}\label{sec:ATL}
To achieve ``Adaption'' and ``Learning'' simultaneously, we propose to augment the environment reward $r_t \in \mathcal{R}^T$ with intrinsic reward $\zeta_t$. By optimizing over mixture of intrinsic and environmental return simultaneously, we achieve transferred policies that try to both follow source advice and learn to acquire skills beyond what source can teach. This trade-off between learning by exploration and learning by adaptation can be realized as follows:
\begin{equation}
    \bar{\eta}_{_{KL, \beta}} = \displaystyle \mathop{\mathbb{E}}_{s_t, a_t\sim \tau}\left(\sum_{t=0}^\infty \gamma^t((1-\beta)r_t - \beta\zeta_t)\right).
    \label{eq:true_objective}
\end{equation}
where the term $\beta$ is the mixing coefficient. The inclusion of intrinsic reward can be seen as adding a directional search to RL exploration. The source optimal policy and source transitions provide a possible direction of search for one step optimal reward at every state along the trajectory $\tau$. 
For consistency of the reward mixing, the rewards $r_t, \zeta_t$ are normalized to form the total reward.

The optimal mixing coefficient $\beta$ is learned over episodic data collected interacting with environment~\cite{li2017metasgd}. The details of hierarchical learning of mixing coefficient is provided in Section-\ref{sec-empirical_learning}, and its behavior in experiments is analyzed in Section-\ref{sec-exp}.
\begin{assumption}
\label{assumption-gaussian_model}
The true transition distribution for source $p^S(.|s,a)$ and the target $p^T(.|s,a)$ are unknown. However, we assume both source and target transition models follow a Gaussian distribution centered at the next propagated state with fixed variance ``$\sigma$'', which is chosen heuristically s.t $$p^S(.|s,a) \sim \mathcal{N}(s'_{t+1}, \sigma), \hspace{4mm} p^T(.|s,a) \sim \mathcal{N}(s_{t+1}, \sigma)$$
\end{assumption}

Although Assumption-\ref{assumption-gaussian_model} might look unrealistic, in reality it is not very restrictive. We empirically show that for all the experimented MuJoCo environments, a bootstrapped Gaussian transition assumption is good enough for ATL agent to transfer the policy efficiently. 

Using the Assumptions-\ref{assumption-source_policy} \& \ref{assumption-gaussian_model}, we can simplify the individual KL term (intrinsic reward) as follows
\begin{equation}
    \zeta_t = \log\left(\pi_{\theta}(a_t|s_t)e^{-(s_{t+1}-s'_{t+1})^2/2\sigma^2}\right).
\end{equation}
The individual terms in the expectation (\ref{eq:adaptive_learining_objective}), $\zeta_t$ represent the distance between two transition likelihoods of landing in the next state $s'_{t+1}$ starting in $s_t$ and under actions $a_t, a'_t$. The target agent is encouraged to take actions that lead to states which are close in expectation to a reference state provided by an optimal baseline policy operating on the source model. The intuition is that by doing so, a possible direction of search for the higher environmental rewards ``$r_t$'' is provided.
\subsection{Actor-Critic}
Using the definition of value function, the objective in (\ref{eq:true_objective}) can be written as $\bar{\eta}_{_{KL, \beta}} = V^{\pi_\theta}(s)$.
The expectation can be rewritten as sum over states and actions as follows:
\begin{equation}
    \bar{\eta}_{_{KL, \beta}} =  \sum_{s \in \mathcal{S}}d^{\pi_\theta}(s)\sum_{a \in \mathcal{A}}\pi_\theta(a|s)Q^{\pi_\theta}(s,a),
    \label{eq:true_objective2}
\end{equation}
where $d^{\pi_\theta}(s)$ is the state visitation distributions \cite{sutton2000policy}.

Considering the case of an off-policy RL, we use a behavioral policy $\psi(a|s)$ for generating trajectories. This behavioral policy is different from the policy $\pi_\theta(a|s)$ to be optimized. The objective function in an off-policy RL measures the total return over the behavioral state visitation distribution, while the mismatch between the training data distribution and the true policy state distribution is compensated by importance sampling estimator as follows,
\begin{equation}
    \bar{\eta}_{_{KL, \beta}} =  \sum_{s \in \mathcal{S}}d^{\psi}(s)\sum_{a \in \mathcal{A}}\left(\psi(a|s)\frac{\pi_\theta(a|s)}{\psi(a|s)}Q^{\psi}(s,a)\right),
    \label{eq:true_objective3}
\end{equation}
Using the previously updated policy as the behavioral policy i.e $\psi(a|s) = \pi_{\theta^-}(a|s)$, the objective expression can be rewritten as,
\begin{equation}
    \bar{\eta}_{_{KL, \beta}} = \displaystyle \mathop{\mathbb{E}}_{s_t \sim d^{\pi_{\theta^-}}, a_t \sim \pi_{\theta^-}}\left(\frac{\pi_\theta(a|s)}{\pi_{\theta^-}(a|s)}\hat{Q}^{\pi_{\theta^{-}}}(s,a)\right).
    \label{eq:true_objective4}
\end{equation}
where $\theta^-$ is the parameter before update and is known to us. An estimated state-action value function $\hat Q$ replaces the true value function $Q$ as the true value function is usually unknown. The mixed reward $r'_{t} = (1-\beta)r_{t} - \beta\zeta_{t}$ is used to compute the critic state-action value function $\hat Q$. Further, any policy update algorithm \cite{sutton2000policy,schulman2017proximal,schulman2015trust, peters2008natural} can be used to update the policy in the direction of optimizing this objective. 

Learning over a mixture of intrinsic and environmental rewards helps in the directional search for maximum total return. The source transitions provide a possible direction in which maximum environmental reward can be achieved.
This trade-off between directional search and random exploration is achieved using the mixed reward. 

\section{Optimization of Target Policy}
In the previous section, we formulated an Adapt-to-Learn policy transfer algorithm; we now describe how to derive a practical algorithm from these theoretical foundations under finite sample counts and arbitrary policy parameterizations. 

Optimizing the following objective (\ref{eq:true_objective4}) we can  generate adaptive policy updates for the target task. 
\begin{eqnarray}
    \pi^{*T}_\theta &=& \arg\max_{\pi_\theta \in \Pi, \beta}\left(\bar{\eta}_{_{KL, \beta}}\right), 
\end{eqnarray}
If calculating the above expectation is feasible, it is possible to maximize the objective in (\ref{eq:true_objective4}) and move the policy parameters in the direction of achieving a higher total discounted return. However, this is not generally the case since the true expectation is intractable, therefore a common practice is to use an empirical estimate of the expectation.

The previous section proposed an optimization method to find the adaptive policy using KL-divergence as an intrinsic reward, enforcing the target transition model to mimic the source transitions. This section describes how this objective can be approximated using a Monte Carlo simulation. The approximate policy update method works by computing an estimate of the gradient of the return and plugging it into a stochastic gradient ascent algorithm  
The gradient estimate over i.i.d data is computed as follows:
\begin{eqnarray}
\hat{g} &=& P_{Z^n}(\nabla_\theta \bar{\eta}_{_{KL, \beta}})\\
&\triangleq& \frac{1}{n}\sum_{i=1}^n\left(\frac{\pi_\theta(a_i|s_i)}{\pi_{\theta^-}(a_i|s_i)}\hat{Q}^{\pi_{\theta^{-}}}(s_i,a_i)\nabla_\theta\log\pi_\theta(a_i|s_i)\right).\nonumber
\end{eqnarray}
where $P_{Z^n}$ is empirical distribution over the data $(Z^n :\{s_i,a_i,a'_i\}_i^n)$.

\begin{algorithm}
    \caption{Adapt-to-Learn Policy Transfer in RL}
    \label{alg:TL}
\begin{algorithmic}[1]
\Require $\pi^{*}(.|s), p^{S}$ \Comment{Source Policy, source simulator}
\State Initialize $s_0^{T} \in \rho_0$. \Comment{Draw initial state from the given distribution in target task}
\For{$i=1 \leq K$}
\While{$s_i \neq terminal$}
\State  $a'_i \sim  \pi^{*}(.|{s}_i)$
\Comment{Generate the optimal action using Source policy}
\State  $a_i \sim  {\pi}_\theta(.|{s}_i)$
\Comment{Generate the action using the ${\pi}_\theta$}
\State $s_{i+1} \sim p^{T}(.|{s}_i,a_i)$
\Comment{Apply the action $a_i$ at state ${s}_i$ in the target task}
\State $s'_{i+1} \sim p^{S}(.|{s}_i,a'_i)$ 
\Comment{Apply the action $a'_i$ at state ${s}_i$ in the source simulator}
\State $\zeta_t = {\pi}_\theta(a_i|s_i)e^{-(s_{i+1}-s'_{i+1})^2/2\sigma^2}$
\Comment{Compute the point-wise KL intrinsic reward}
\State $Z_i = (\{s_i,r_i,\zeta_i,a_i, a'_i\})$
\Comment{Incrementally store the trajectory for policy update}
\EndWhile
\State $P_{Z_{train}^n}(\bar{\eta}_{_{KL, \beta}})$
\Comment{Form the Empirical loss}
\State $\theta' \leftarrow \theta + \alpha P_{Z_{train}^n}\left(\nabla_\theta\bar{\eta}_{_{KL,\beta}}\right) $
\Comment{Maximize the total return to update the policy}
\State Collect test trajectories $Z_{test}^n$ using $\pi_{\theta'}$
\State $\beta \leftarrow \beta + \bar{\alpha} P_{Z_{test}^n}\left(\nabla_{\beta}\bar{\eta}_{_{KL,\beta}}\right)$
\Comment{Maximize the total return for optimal mixing coefficient}
\EndFor
\end{algorithmic}
\end{algorithm}
\subsection{Sample-Based Estimation of the Gradient}
\label{sec-empirical_learning}
The previous section proposed an optimization method to find the adaptive policy using KL-divergence as an intrinsic reward, enforcing the target transition model to mimic the source transitions. This section describes how this objective can be approximated using a Monte Carlo simulation. The approximate policy update method works by computing an estimate of the gradient of the return and plugging it into a stochastic gradient ascent algorithm  
The gradient estimate over i.i.d data from the collected trajectories is computed as follows:
\begin{eqnarray}
\hat{g} &=& P_{Z^n}(\nabla_\theta \bar{\eta}_{_{KL, \beta}})\\
&\triangleq& \frac{1}{n}\sum_{i=1}^n\left(\frac{\pi_\theta(a_i|s_i)}{\pi_{\theta^-}(a_i|s_i)}\hat{Q}^{\pi_{\theta^{-}}}(s_i,a_i)\nabla_\theta\log\pi_\theta(a_i|s_i)\right) \nonumber.
\end{eqnarray}
where $P_{Z^n}$ is empirical distribution over the data $(Z^n :\{s_i,a_i,a'_i\}_i^n)$. 
The details of empirical estimate of the gradient $\hat{g}$ of the total return $\bar{\eta}_{_{KL, \beta}}$ is provided in Appendix.

Note that the importance sampling based off-policy update objective renders our discounted total return and its gradient independent of policy parameter $\theta$. Hence the gradient of state-action value estimates $\nabla_\theta \hat{Q}^{\pi_{\theta^{-}}}(s,a)$ with respect to $\theta$ is zero in the above expression for total gradient.

\subsection{Learning Mixing Coefficient from Data}
A hierarchical update of the mixing coefficient $\beta$ is carried out over  n-test trajectories, collected using the updated policy network $\pi_{\theta'}(a|s)$. Where $\theta'$ is parameter after the policy update step. We use stochastic gradient ascent to update the mixing coefficient $\beta$ as follows
$$
\beta \leftarrow \beta + \bar{\alpha} \hat{g}_\beta,
\hspace{2mm}
s.t \hspace{2mm} 0 \leq \beta \leq 1 .
$$
where $\bar{\alpha}$ is the learning rate and $\hat{g}_\beta = P_{Z^n_{test}}(\nabla_\beta \bar{\eta}_{_{KL, \beta}})$ is the empirical estimate of the gradient of the total return $\bar{\eta}_{_{KL, \beta}}(\pi_{\theta'}, \pi^*)$. 

A hierarchical update of the mixing coefficient $\beta$ is carried out over  n-test trajectories, collected using the updated policy network $\pi_{\theta'}(a|s)$. 
The mixing coefficient $\beta$ is learnt by optimizing the return over trajectory as follows,
$$
\beta = arg\max_\beta(\bar{\eta}_{KL,\beta}(\pi_{\theta'}, \pi^*))
$$
where $\theta'$ is parameter after the policy update step. 
$$
\beta = arg\max_\beta\mathop{\mathbb{E}}_{s_t,a_t \sim \tau}\left(p_{\pi_\theta}(\tau)\sum_{t=1}^\infty \gamma^tr'_t\right)
$$
We can use gradient ascent to update parameter $\beta$ in direction of optimizing the reward mixing as follows,
$$
\beta \leftarrow \beta+\bar{\alpha}\nabla_\beta(\bar{\eta}_{KL,\beta}(\pi_{\theta'}, \pi^*)).
$$
Using the definition of mixed reward as $r'_t = (1-\beta)r_t - \beta\zeta_t$, we can simplify the above gradient as,
$$
\beta \leftarrow \beta+\bar{\alpha}\mathop{\mathbb{E}}_{s_t,a_t \sim \tau}\left(p_{\pi_\theta}(\tau)\sum_{t=1}^\infty \gamma^t\nabla_\beta(r'_t)\right)
$$
$$
\beta \leftarrow \beta+\bar{\alpha}\mathop{\mathbb{E}}_{s_t,a_t \sim \tau}\left(\sum_{t=1}^\infty \gamma^t(r_t - \zeta_t)\right).
$$
We use stochastic gradient ascent to update the mixing coefficient $\beta$ as follows
$$
\beta \leftarrow \beta + \bar{\alpha} \hat{g}_\beta,
\hspace{2mm}
s.t \hspace{2mm} 0 \leq \beta \leq 1 .
$$
where $\bar{\alpha}$ is the learning rate and $\hat{g}_\beta = P_{Z^n_{test}}(\nabla_\beta \bar{\eta}_{_{KL, \beta}})$ is the empirical estimate of the gradient of the total return $\bar{\eta}_{_{KL, \beta}}(\pi_{\theta'}, \pi^*)$. The gradient estimate $\hat{g}_\beta$ over data $(Z^n_{test} :\{s_i,a_i,a'_i\}_i^T)$ is computed as follows,
$$
\hat{g}_\beta = \frac{1}{N}\sum_{i=1}^N\left(\sum_{t=1}^\mathcal{H} \gamma^t(r_t - \zeta_t)\right)
$$
where $\mathcal{H}$ truncated trajectory length from experiments.

As we can see the gradient of objective with respect to mixing coefficient $\beta$ is an average over difference between environmental and intrinsic rewards. If $r_t-\zeta_t \geq 0$ the update will move parameter $\beta$ towards favoring learning through exploration more than learning through adaptation and visa versa.

\begin{figure}[tbh!]
    \centering
    \includegraphics[width=1\columnwidth]{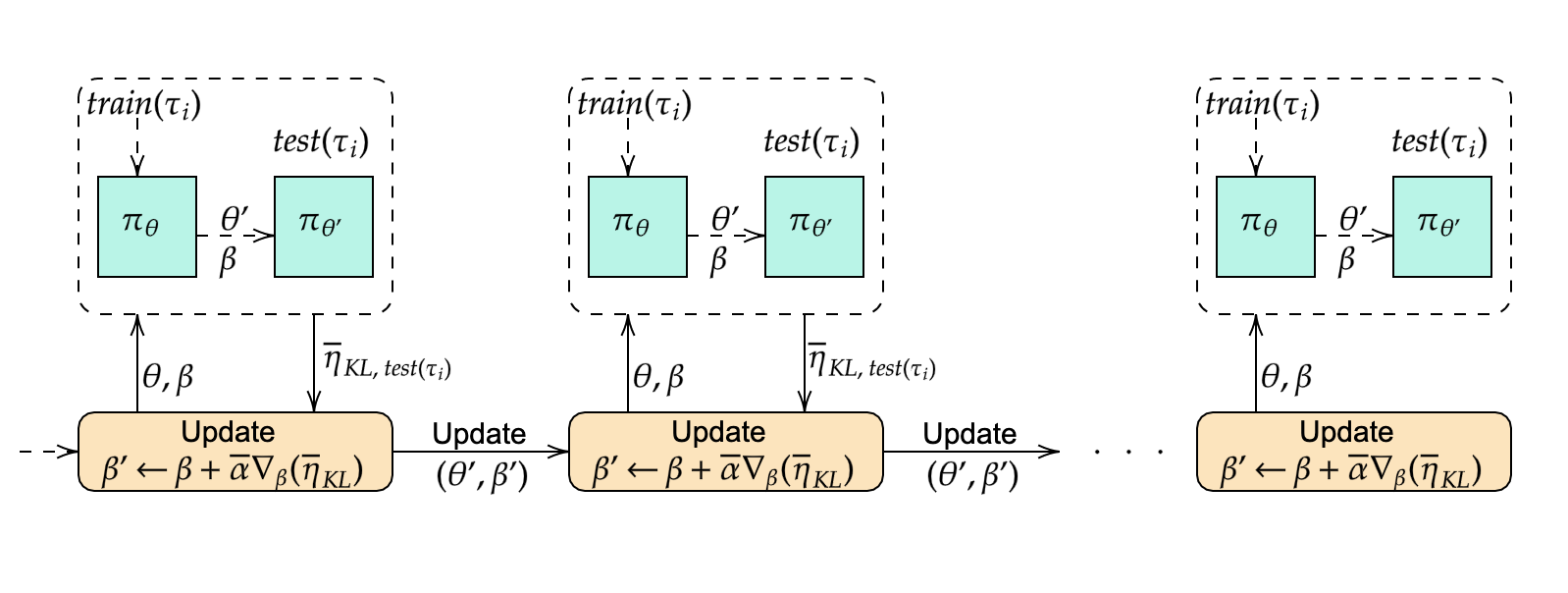}
    \caption{Total Policy Update scheme to learn Target task. The Hyper-parameter $\beta$ are updated over test trajectories generated using the updated policy parameters.}
    \label{fig:meta-sgd}
\end{figure}

As $\beta$ update is a constrained optimization with constraint $0 \leq \beta \leq 1$. We handle this constrained optimization by modelling as output of sigmoidal network $\beta = \sigma(\phi)$ parameterized by parameters $\phi$. The constrained optimization can be equivalently written as optimizing w.r.to $\phi$ as follows
$$
\phi \leftarrow \phi + \bar{\alpha} \hat{g}_\beta \nabla_\phi(\beta),
\hspace{2mm}
where \hspace{2mm} \beta = \sigma(\phi)
$$
The details of policy and $\beta$ update is shown in Figure-\ref{fig:meta-sgd}.
\begin{table}[tbh!]
\centering
    \begin{tabular}{|c|c|c|c|c|}
    \hline
     \textbf{Env}&\textbf{Property}&\textbf{Source}&\textbf{Target}&\%\textbf{Change}\\
     \hline
     Hopper&Floor Friction&1.0&2.0&+100\%\\
     \hline
     HalfCheetah&gravity&-9.81&-15&+52\%\\
     &Total Mass&14&35&+150\%\\
     &Back-Foot Damping&3.0&1.5&-100\%\\
     &Floor Friction&0.4&0.1&-75\%\\
     \hline
     Walker2d&Density&1000&1500&+50\%\\
     &Right-Foot Friction&0.9&0.45&-50\%\\
     &Left-Foot Friction&1.9&1.0&-47.37\%\\
     \hline
\end{tabular}
    \caption{Transition Model and environment properties for Source and Target task and \% change}
    \label{Table:env_property}
\end{table}
\begin{table}[tbh!]
\centering
    \begin{tabular}{|c|ccc|}
         \hline
            & \textbf{Hopper}&\textbf{Walker2d}& \textbf{HalfCheetah} \\
         \textbf{State Space}& 12&18&17\\
         \textbf{Action Space}& 3&6&6\\
         \textbf{Number of layers}&3&3&3\\
         \textbf{Layer Activation}&tanh&tanh&tanh\\
         \textbf{Network Parameters}&10530&28320 &26250\\
         \hline
         \textbf{Discount}&0.995&0.995&0.995\\
         \textbf{Learning rate} ($\alpha$)&$1.5\times10^{-5}$&$8.7\times10^{-6}$&$9\times10^{-6}$\\
         $\beta$ \textbf{Initial Value}&$0.5$&$0.5$&$0.5$\\
         $\beta$-\textbf{Learning rate} ($\bar{\alpha}$)&$0.1$&$0.1$&$0.1$\\
         \textbf{Batch size}&20&20&5\\
         \textbf{Policy Iter}&3000&5000&1500\\
         \hline
    \end{tabular}
    \caption{Policy Network details and Network learning parameter details}
    \label{tab:network_and Learning_details}
\end{table}
\section{Sample Complexity and $\epsilon$-Optimality}
In this section we will provide without proofs the sample complexity and $\epsilon$-optimality results of the proposed policy transfer method. The proofs of the following theorem can be found in.
\subsection{Lower bounds on Sample Complexity}
Although there is some empirical evidence that transfer can improve performance in subsequent reinforcement-learning tasks, there are not many theoretical guarantees. Since many of the existing transfer algorithms approach the problem of transfer as a method of providing good initialization to target task RL, we can expect the sample complexity of those algorithms to still be a function of the cardinality of state-action pairs $|N| = |\mathcal{S}| \times |\mathcal{A}|$. On the other hand, in a supervised learning setting, the theoretical guarantees of the most algorithms have no dependency on size (or dimensionality) of the input domain (which is analogous to $|N|$ in RL). Having formulated a policy transfer algorithm using labeled reference trajectories derived from optimal source policy, we construct supervised learning like PAC property of the proposed method. For deriving, the lower bound on the sample complexity of the proposed transfer problem, we consider only the adaptation part of the learning i.e., the case when $\beta = 1$ in Eq-(\ref{eq:true_objective}). This is because, in ATL, adaptive learning is akin to supervised learning, since the source reference trajectories provide the target states 

Suppose we are given the learning problem specified with training set $Z^n = (Z1, \ldots Z_n)$ where each $Z_i = (\{s_i, a_i\})_{i=0}^n$ are  independently drawn trajectories according to some distribution $P$. Given the data $Z^n$ we can compute the empirical return $P_{Z^n}(\bar{\eta}_{_{KL,\beta}})$ for every $\pi_\theta \in \Pi$,  we will show that the following holds:
\begin{equation}
    \|P_{Z^n}(\bar{\eta}_{_{KL,\beta}}) - P(\bar{\eta}_{_{KL,\beta}})\| \leq \epsilon.
\end{equation}
with probability at least $1-\delta$, for some very small $\delta$ s.t $0 \leq \delta \leq 1$.  We can claim that the empirical return for all $\pi_\theta$ is a sufficiently accurate estimate of the true return function. Thus a reasonable learning strategy is to  find a $\pi_\theta \in \Pi$ that would minimize empirical estimate of the objective Eq-(\ref{eq:true_objective}).

\begin{theorem}
If the induced class of the policy $\pi_\theta$:$\mathcal{L}_\Pi$ has uniform convergence property in empirical mean; then the empirical risk minimization is PAC.
s.t
\begin{equation}
    P^n( P(\bar{\eta}_{_{KL, \hat{\pi}^{*}}}) - P(\bar{\eta}_{_{KL, {\pi^*}}}) \geq \epsilon) \leq \delta .
    \label{eq:12}
\end{equation}
and number of trajectory samples required can be lower bounded as
\begin{equation}
    n(\epsilon,\delta) \geq \frac{2}{\epsilon^2(1-\gamma)^2}\log\left(\frac{2|\Pi|}{\delta}\right).
\end{equation}
\end{theorem}
Please refer Appendix for the proof of the above theorem.
\subsection{$\epsilon$-Optimality result under Adaptive Transfer-Learning}
Consider MDP $M^*$ and $\hat{M}$ which differ in their transition models. For the sake of analysis,
let $M^*$ be the MDP with ideal transition model, such that target follows source transition $p^*$ precisely. Let $\hat{p}$ be the transition model achieved using the estimated policy learned over data interacting with the target model and the associated MDP be denoted as $\hat{M}$.  We analyze the $\epsilon$-optimality of return under adapted source optimal policy through ATL.

\begin{definition}
\label{defn-1}
Given the value function $V^* = V^{\pi^*}$ and model $M_1$ and $M_2$, which only differ in the corresponding transition models $p_1$ and $p_2$. Define $\forall s,a \in \mathcal{S} \times \mathcal{A}$
$$d_{M_1,M_2}^{V^*} = \sup_{s,a \in \mathcal{S} \times \mathcal{A}}\left\vert \displaystyle \mathop{\mathbb{E}}_{s' \sim P_1(s,a)} [V^{*}(s')] - \displaystyle \mathop{\mathbb{E}}_{s' \sim {P}_2(s,a)} [V^{*}(s')]\right\vert.$$
\end{definition}
\vspace{3mm}
\begin{lemma}
\label{lemma:epsilon-optimality}
Given $M^*$, $\hat M$ and value function $V^{\pi^*}_{M^*}$, $V^{\pi^*}_{\hat M}$ the following bound holds 
$\left\Vert V^{\pi^*}_{M^*} -  V^{\pi^*}_{\hat M}\right\Vert_{\infty} \leq \frac{\gamma \epsilon}{(1-\gamma)^2}$
\end{lemma}
where $\max_{s,a}\|\hat{p}(.|s,a)-p^*(.|s,a)\| \leq \epsilon$ and $\hat{p}$ and $p^*$ are transition of MDP $\hat{M}, M^*$ respectively.

The proof of this lemma is based on the simulation lemma \cite{kearns2002near}. For the proof of above lemma refer Appendix. 
Similar results for RL with imperfect models were reported by \cite{NanJiang}. 
\section{Policy transfer in simulated robotic locomotion tasks}\label{sec-exp}
\begin{figure*}[tbh!]
	\centering
	\begin{subfigure}{0.25\linewidth}
		\includegraphics[width=\linewidth]{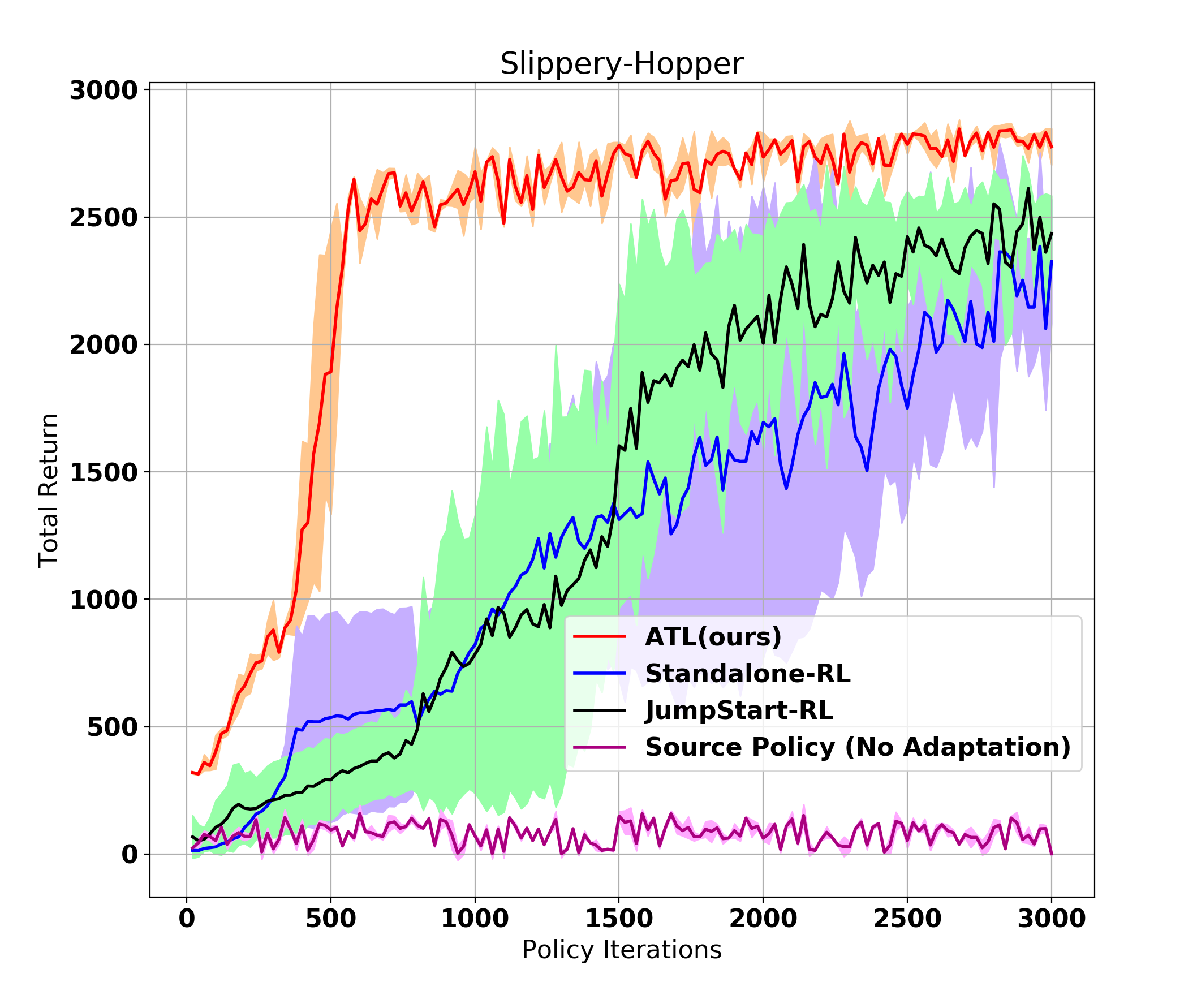}
		\caption{}
		\label{fig:env_Hopper}
	\end{subfigure}
	\begin{subfigure}{0.25\linewidth}
		\includegraphics[width=\textwidth]{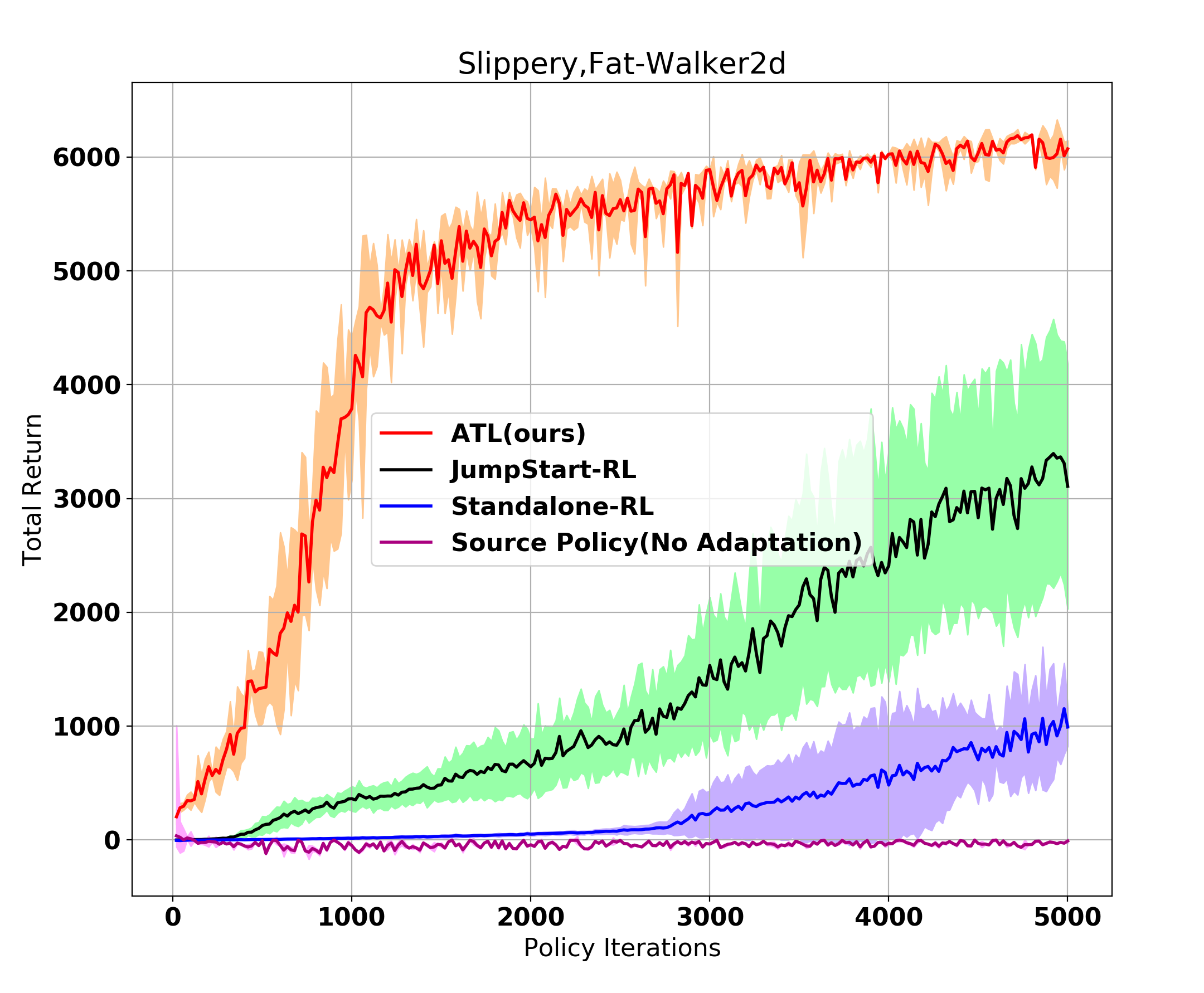}
		\caption{}
		\label{fig:env_Walker2d}
	\end{subfigure}
    \begin{subfigure}{0.25\linewidth}
		\includegraphics[width=\textwidth]{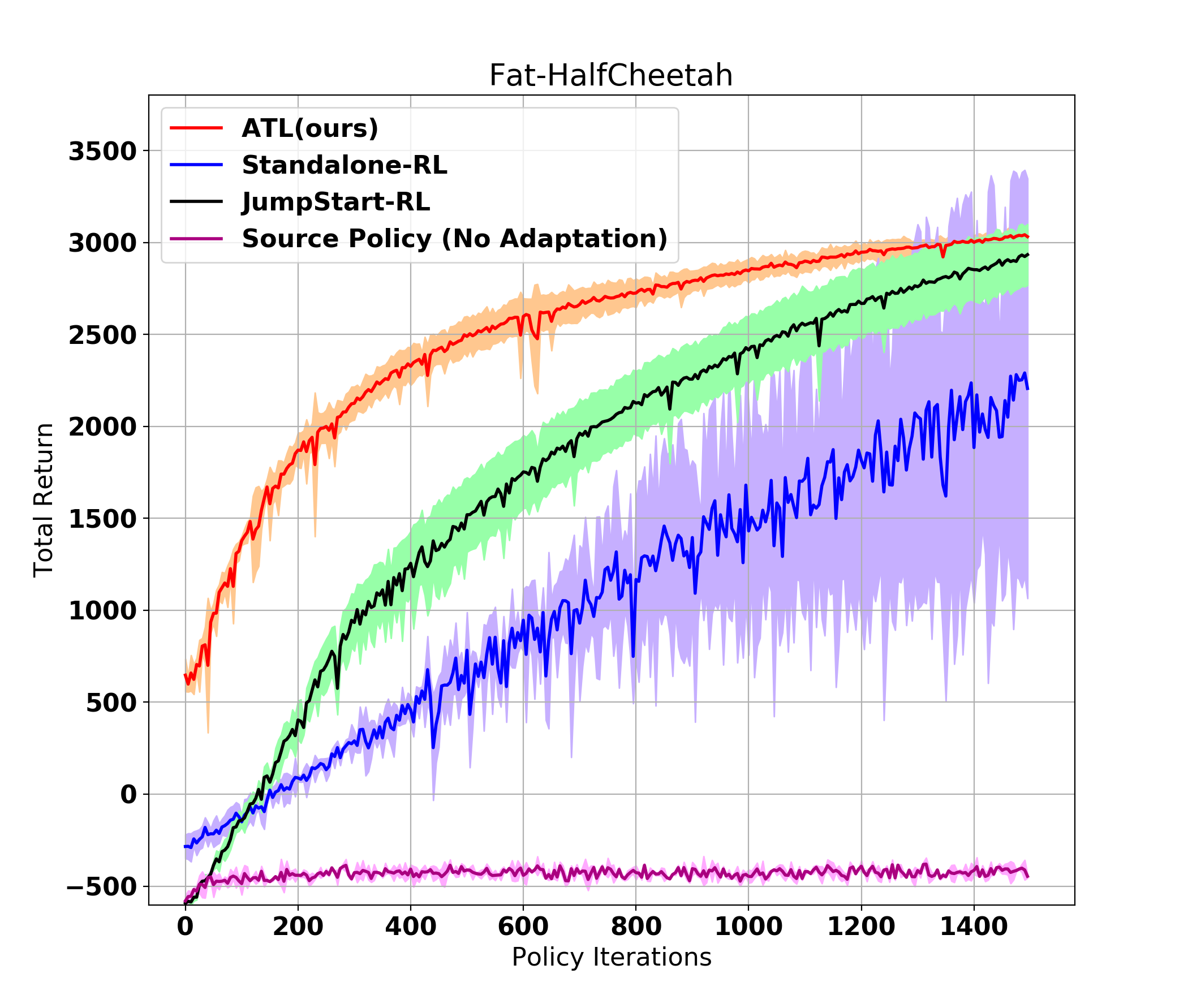}
		\caption{}
		\label{fig:env_HalfCheetah}
	\end{subfigure}
	\caption{Learning curves for locomotion tasks, averaged across five runs of each algorithm: Adapt-to-Learn(Ours), Randomly Initialized RL(PPO), Warm-Started PPO using source policy parameters and Best case imitation learning using Source policy directly on Target Task without any adaptation.}
	\label{fig:Env_results}
\end{figure*}
\begin{figure*}[tbh!]
	\centering
	\begin{subfigure}{0.25\linewidth}
		\includegraphics[width=\textwidth]{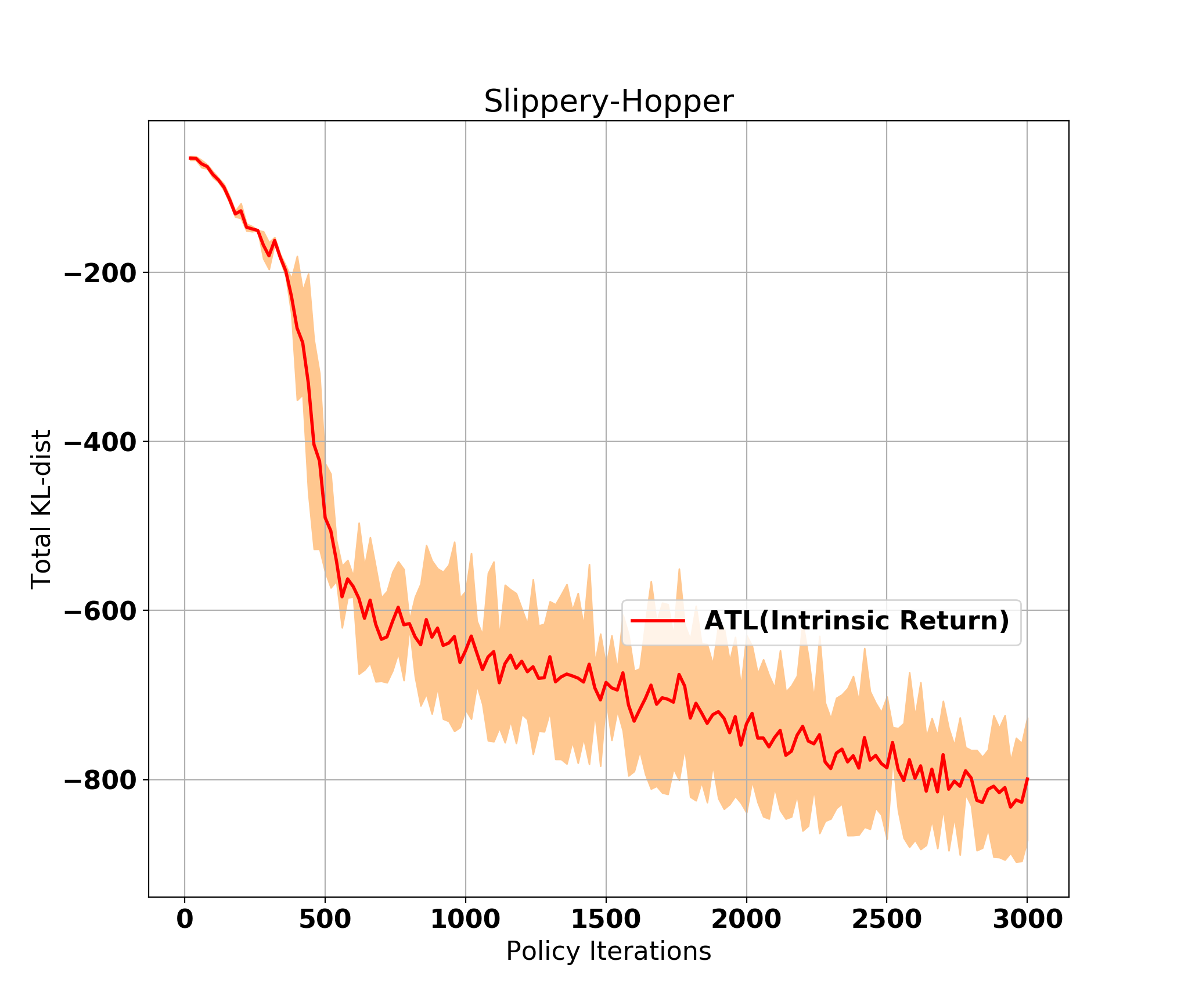}
		\caption{}
		\label{fig:Quad_trajectory}
	\end{subfigure}
	\begin{subfigure}{0.25\linewidth}
		\includegraphics[width=\textwidth]{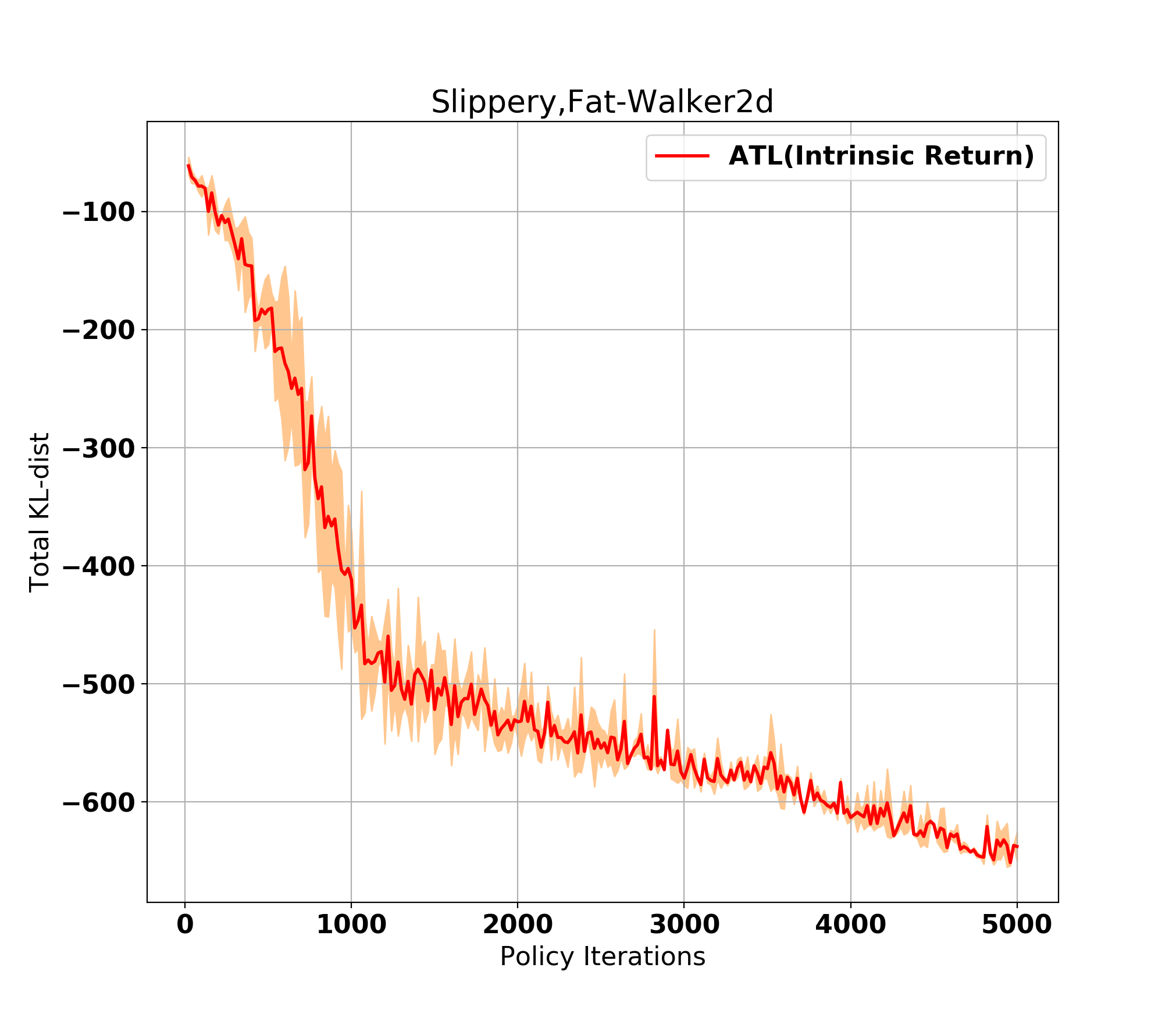}
		\caption{}
		\label{fig:State_tracking}
	\end{subfigure}
    \begin{subfigure}{0.25\linewidth}
		\includegraphics[width=\textwidth]{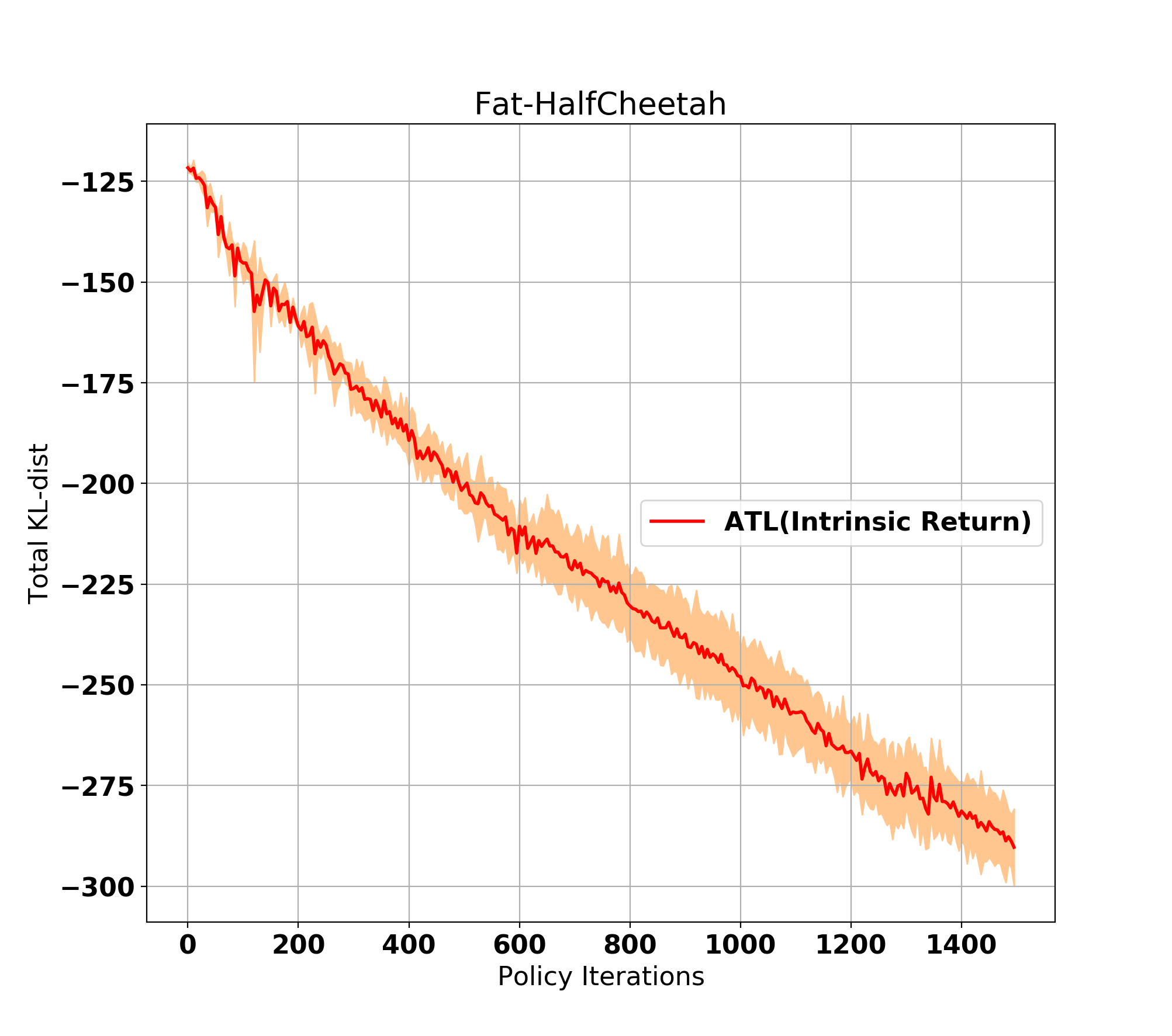}
		\caption{}
		\label{fig:Quad_net_perf}
	\end{subfigure}
	\caption{Trajectory KL divergence Total Intrinsic Return $\left(-\sum e^{\zeta_t}\right)$ averaged across five runs.}
	\label{fig:Env_KL}
\end{figure*}
\begin{figure}[tbh!]
    \centering
    \includegraphics[width=\columnwidth]{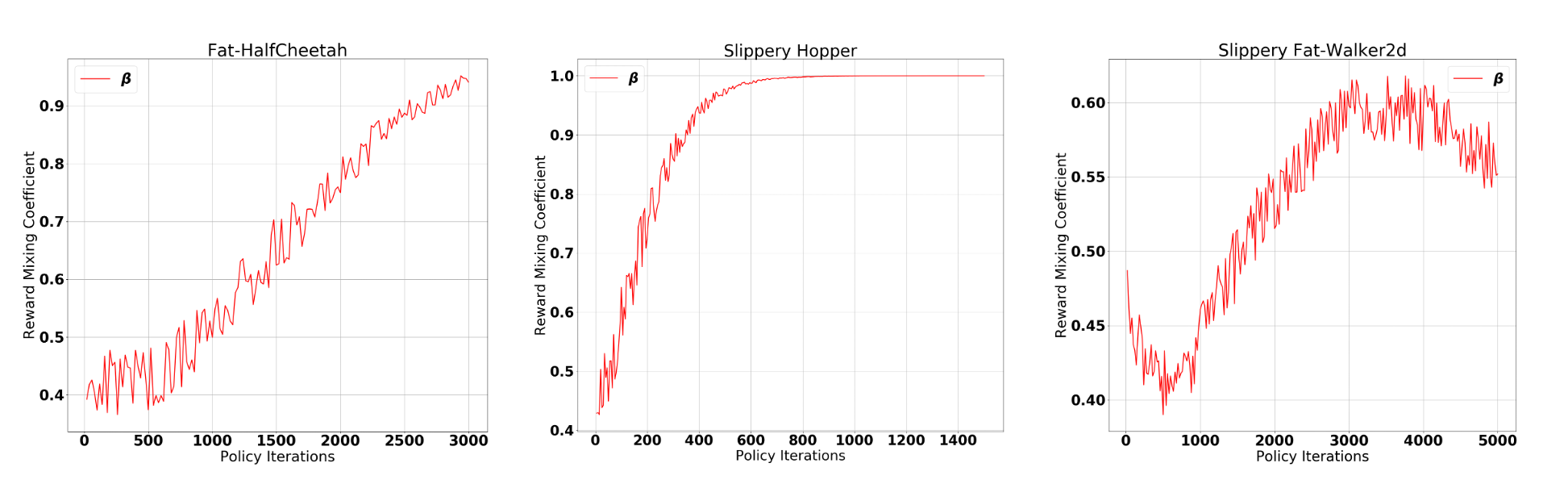}
    \caption{Mixing coefficient $\beta$ over one run for HalfCheetah, Hopper and Walker2d Environment}
    \label{fig:beta_plot}
\end{figure}
\begin{figure*}[tbh!]
	\centering
	\begin{subfigure}{0.25\linewidth}
		\includegraphics[width=\textwidth]{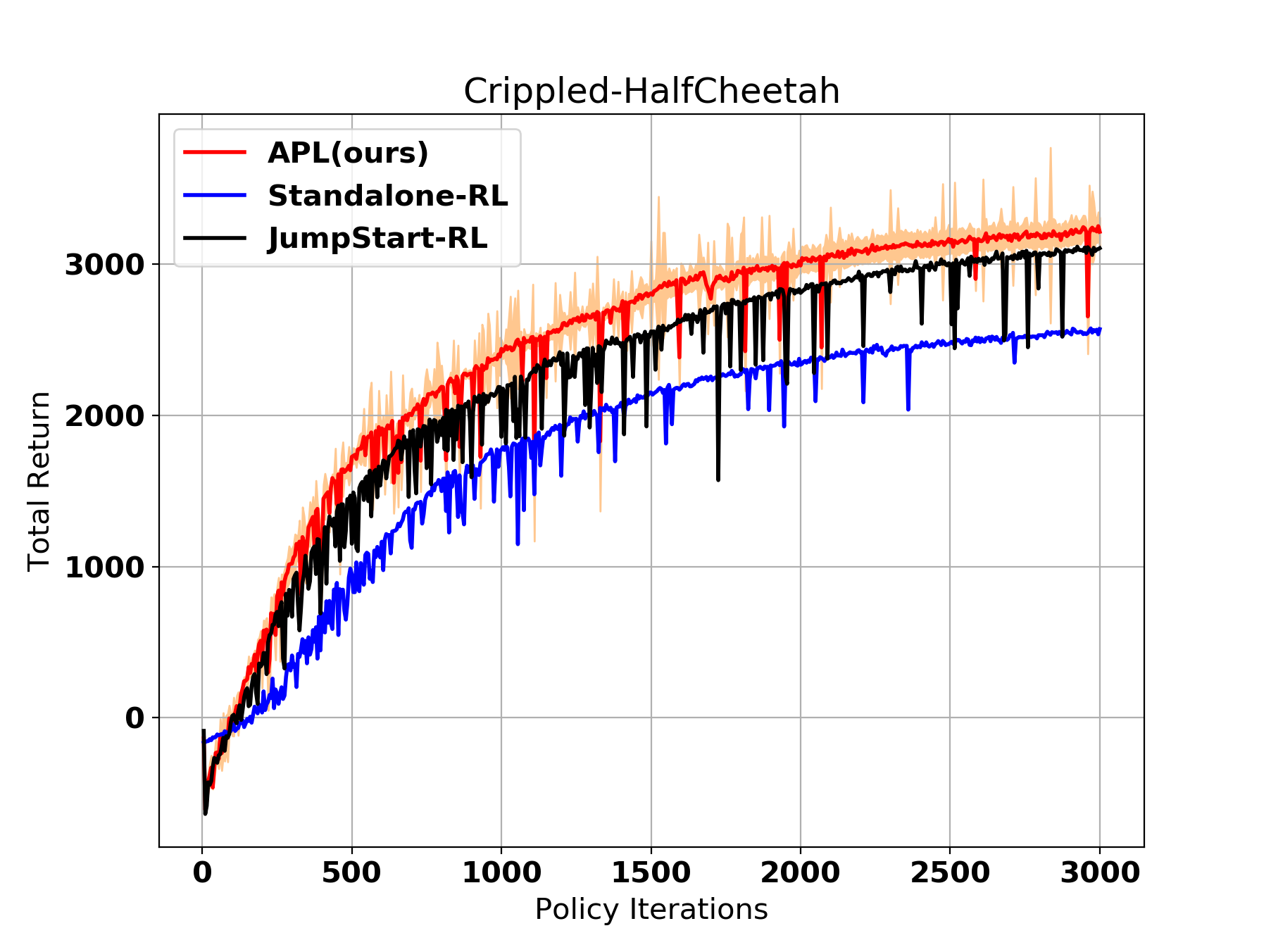}
		\caption{}
		\label{fig:Quad_trajectory}
	\end{subfigure}
	\begin{subfigure}{0.25\linewidth}
		\includegraphics[width=\textwidth]{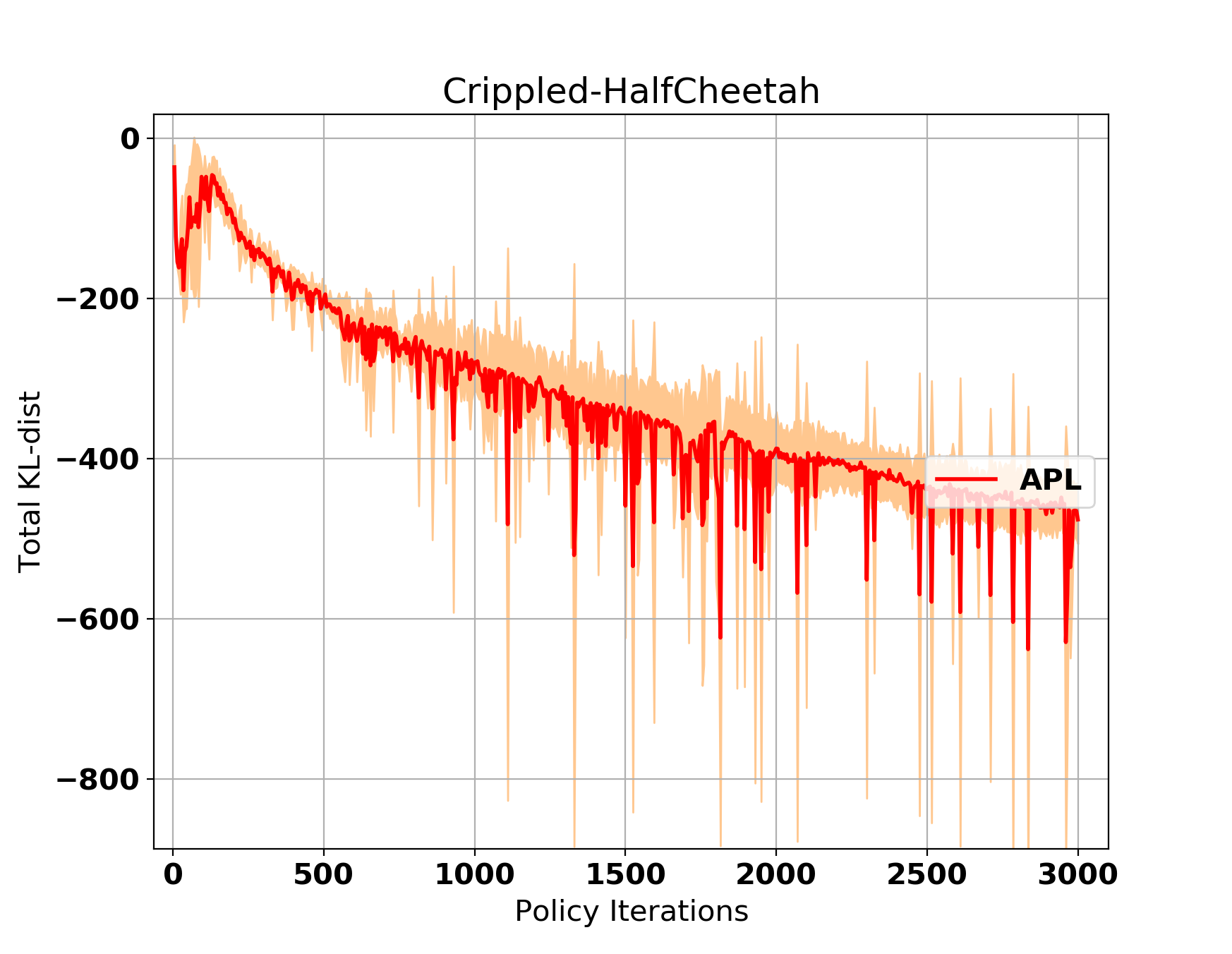}
		\caption{}
		\label{fig:State_tracking}
	\end{subfigure}
    \begin{subfigure}{0.25\linewidth}
		\includegraphics[width=\textwidth]{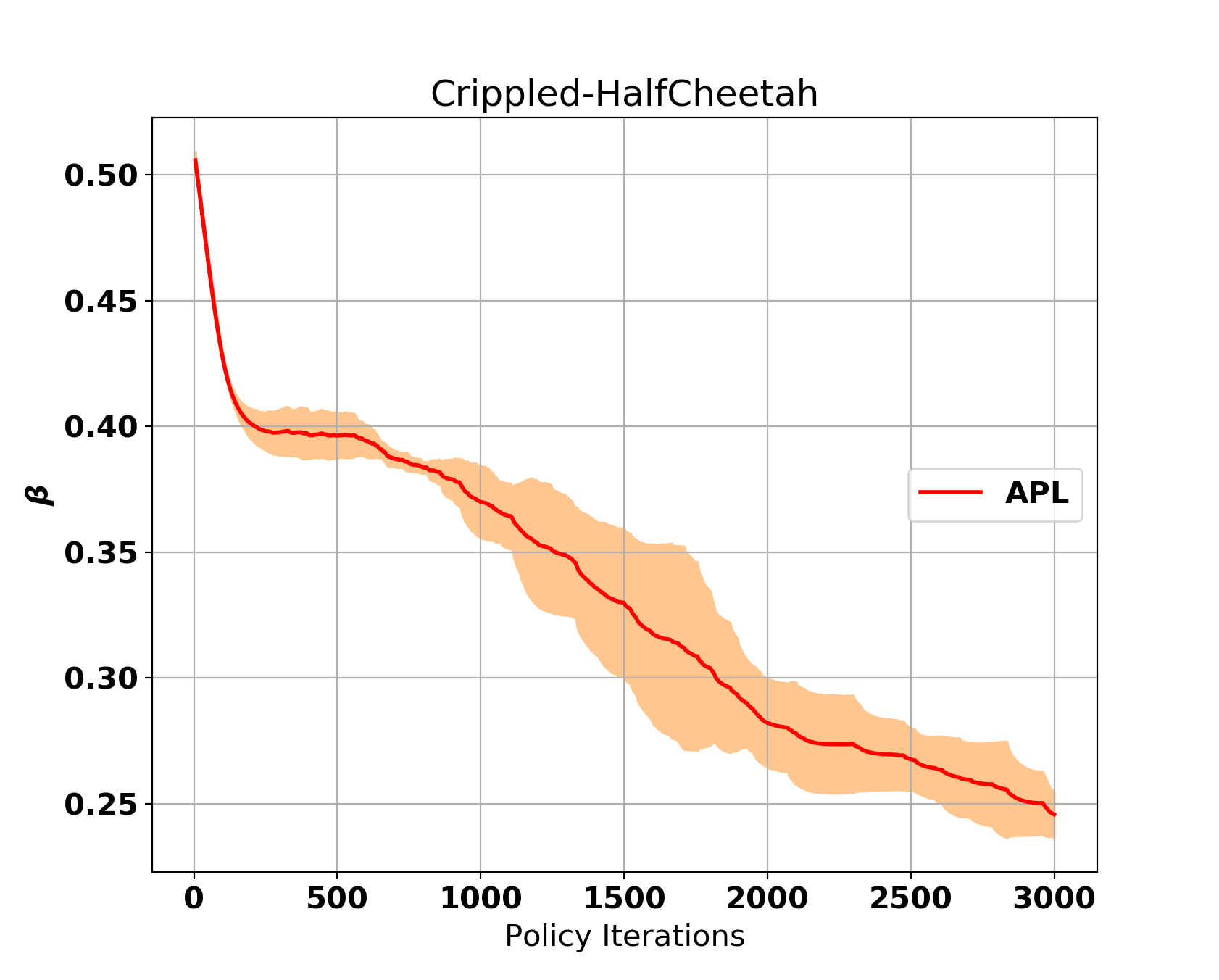}
		\caption{}
		\label{fig:Quad_net_perf}
	\end{subfigure}
	\caption{(a) Learning curves for Crippled-Fat HalfCheetah Env, averaged across five runs: Adapt-to-Learn (ATL (Ours)), Randomly Initialized RL (PPO), Warm-Started PPO using source policy parameters. (b) Avg Intrinsic reward for ATL. (c) Mixing coefficient $\beta$.}
	\label{fig:crippled_hc}
\end{figure*}
To evaluate Adapt-to-Learn Policy Transfer in reinforcement learning, we design our experiments using sets of tasks based on the continuous control environments in MuJoCo simulator \cite{todorov2012mujoco}.
Our experimental results demonstrate that ATL can adapt to significant changes in transition dynamics. We perturb the parameters of the simulated target models for the policy transfer experiments (see Table-\ref{Table:env_property} for original and perturbed parameters of the target model). To create a challenging training environment, we changed the parameters of the transition model such that the optimal source policy alone without further learning cannot produce any stable results (see source policy performance in Figure-\ref{fig:Env_results} \& \ref{fig:crippled_hc}). We compare our results against two baselines: (a) Initialized RL (Initialized PPO) (Warm-Start-RL \cite{ammar2015unsupervised}) (b) Stand-alone reinforcement policy learning (PPO) \cite{schulman2017proximal}. 

We experiment with the ATL algorithm on Hopper, Walker2d, and HalfCheetah Environments.
The states of the robots are their generalized positions and velocities, and the actions are joint torques. High dimensionality, non-smooth dynamics due to contact discontinuity, and under-actuated dynamics make these tasks very challenging. We use deep neural networks to represent the source and target policy, the details of which are in the Table-\ref{tab:network_and Learning_details}.

Learning curves showing the total reward averaged across five runs of each algorithm are provided in Figure-\ref{fig:Env_results} \& \ref{fig:crippled_hc}. Adapt-to Learn policy transfer solved all the tasks, yielding quicker learning compared to other baseline methods. 
Figure-\ref{fig:beta_plot} provide the beta vs episode plot for HalfCheetah, Hopper and Walker2d environment. Recall that the extent of mixing of adaptation and learning from exploration is driven by the mixing coefficient $\beta$. Figure-\ref{fig:beta_plot} shows that for all these environments with parameter perturbation, the ATL agent found it valuable to follow source baseline and therefore favored adaptation over exploration in search for optimal policy. In cases where source policy is not good enough to learn the target task the proposed algorithm seamlessly switches to initialized RL by favouring the exploration, as demonstrated in Crippled Half Cheetah experiment: 

To further test the algorithm's capability to adapt, we evaluate its performance in the crippled-Halfcheetah environment (Figure-\ref{fig:crippled_hc}). With a loss of actuation in the front leg, adaptation using source policy might not aid learning, and therefore learning from exploration is necessary. We observe the mixing coefficient $\beta$ evolves  closer to zero, indicating that ATL is learning to rely more on exploration rather than following the source policy. This results in a slight performance drop initially, but overall, ATL outperforms the baselines.  This shows that in the scenarios where adaptation using source policy does not solve the task efficiently, ATL learns to seamlessly switch to exploring using environmental rewards. 

\section{Conclusion}
We introduced a new transfer learning technique for RL: Adapt-to-Learn (ATL), that utilizes combined adaptation and learning for policy transfer from source to target tasks. We demonstrated on nonlinear and continuous robotic locomotion tasks that our method  leads to a significant reduction in sample complexity over the prevalent warm-start based approaches. 
We demonstrated that ATL seamlessly learns to rely on environment rewards when learning from Source does not provide a direct benefit, such as in the Crippled Cheetah environment. There are many exciting directions for future work. A network of policies that can generalize across multiple tasks could be learned based on each new adapted policies. How to train this end-to-end is an important question. The ability of Adapt-to-Learn to handle significant perturbations to the transition model indicates that it should naturally extend to sim-to-real transfer \cite{peng2017sim,ross2011reduction,yan2017sim}, and cross-domain transfer \cite{wang2009manifold}. 
Another exciting direction is to extend the work to other combinatorial domains (e.g., multiplayer games). We expect, therefore, follow on work will find other exciting ways of exploiting such adaptation in RL, especially for robotics and real-world applications. 


\bibliographystyle{unsrt}
\bibliography{main}
\appendix
\subsection{Related work}  
Deep Reinforcement Learning (D-RL) has recently enabled agents to learn policies for complex robotic tasks in simulation \cite{peng2016terrain,peng2017deeploco,liu2017learning,heess2017emergence}. However, D-RL has been plagued by the curse of sample complexity.  Therefore, the capabilities demonstrated in the simulated environment are hard to replicate in the real world. 
This learning inefficiency of D-RL has led to significant work in the field of Transfer Learning (TL) \cite{taylor2009transfer}. A significant body of literature on transfer in RL is focused on initialized RL in the target domain using learned source policy; known as jump-start/warm-start methods \cite{taylor2005value,ammar2012reinforcement,ammar2015unsupervised}. Some examples of these transfer architectures include transfer between similar tasks \cite{banerjee2007general}, transfer from human demonstrations \cite{peters2006policy} and transfer from simulation to real \cite{peng2017sim,ross2011reduction,yan2017sim}.
Efforts have also been made in exploring accelerated learning directly on real robots, through Guided Policy Search (GPS) \cite{levine2015learning} and parallelizing the training across multiple agents using meta-learning \cite{levine2016learning,nagabandi2018neural,zhu2018dexterous}. Sim-to-Real transfers have been widely adopted in the recent works and can be viewed as a subset of same domain transfer problems. Daftry et al. \cite{daftry2016learning} demonstrated the policy transfer for control of aerial vehicles across different vehicle models and environments. Policy transfer from simulation to real using an inverse dynamics model estimated interacting with the real robot is presented by \cite{christiano2016transfer}.
The agents trained to achieve robust policies across various environments through learning over an adversarial loss is presented in  \cite{wulfmeier2017mutual}.  

\subsection{Derivation of Behavioral Adaptation KL Divergence Intrinsic Reward}
\begin{figure*}[tbh!]
	\centering
	\begin{subfigure}{0.8\columnwidth}
		\includegraphics[width=\linewidth]{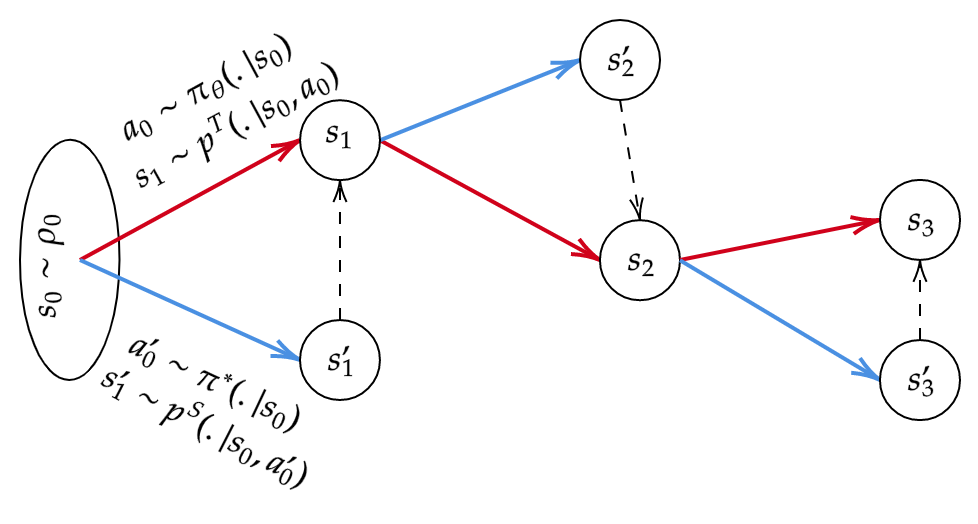}
		\caption{}
		\label{fig:Trajectory}
	\end{subfigure}
	\begin{subfigure}{0.8\columnwidth}
		\includegraphics[width=\textwidth]{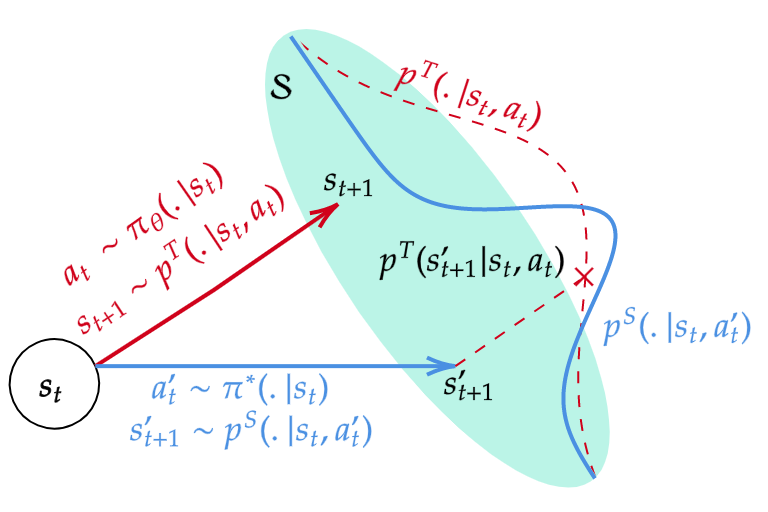}
		\caption{}
		\label{fig:One_step_traj}
	\end{subfigure}
	\caption{(a) Target Trajectory under policy $\pi_\theta$ and local trajectory deviation under source optimal policy $\pi^*$ and source transition $p^S$ (b) One step Target and Source transition(simulated) starting from state $s_t$. Transition likelihood $p^T(s'_{t+1}|s_t, a_t)$ is the probability of landing in state $s'_{t+1}$ starting from state $s_t$ and using action target $a_t$ under target transition model.}
	\label{fig:Traj_deviation}
\end{figure*}
The adaptation objective can be formalized as minimizing the average KL-divergence \cite{schulman2015trust} between source and target transition trajectories as follows,
\begin{eqnarray}
\eta_{_{KL}}(\pi_\theta, \pi^*) &=& D_{KL}(p_{\pi_\theta}(\tau)\|q_{\pi^*}(\tau)), \nonumber \\
    \eta_{_{KL}}(\pi_\theta, \pi^*) &=& \int_{\mathcal{S},\mathcal{A}} p_{\pi_\theta}(\tau)\log\left(\frac{p_{\pi_\theta}(\tau)}{q_{\pi^*}(\tau)}\right)d\tau
    \label{eq:KL_definition}
\end{eqnarray}
where $\tau = (s_0, s_1, s_2,\dots)$ is the trajectory in the target domain under the policy $\pi_{\theta}(.|s)$ defined as collection states visited starting from state $s_0 \sim \rho_0$ and making transitions under target transition model $p^T(.|s_t,a_t)$.

In the above defined KL divergence term the random variable is the trajectory $\tau = (s_0, s_1, s_2,\dots)$. We explain the flow of the algorithm in Figure \ref{fig:Traj_deviation}. The algorithm starts with some random state $s_0 \sim \rho_0$ and using source optimal policy $\pi_\theta(.|s_0)$ and target transition model $p^T(.|s_0,\pi_\theta(s_0))$, make a transition to state $s_1$ (Red arrow, Figure-\ref{fig:Traj_deviation}a). The source simulator is now initialized to state $s_0$ and using source optimal policy $\pi^*(.|s_0)$ we make optimal transition under source transition model $p^S(.|s_0, \pi^*(s_0))$ to state $s'_1$ (Blue arrow, this is a simulated step using source simulator Figure-\ref{fig:Traj_deviation}a). 

The likelihood of landing in the reference state $s'_1$( obtained from optimal source transition) is now evaluated, under target transition model and target policy. We call this likelihood as trajectory deviation likelihood. The trajectory deviation likelihood can be expressed as $p^T(s'_1|s_0,\pi_\theta(s_0))$. Note that we obtain this likelihood by evaluation the target transition probability at state $s'_1$. For the next step of learning the source model is reinitialized to the target transitioned state $s_1$ (dotted arrow in Figure-\ref{fig:Traj_deviation}b and the process is repeated as above. 

One step KL divergence between transition probabilities or one-step Intrinsic reward can be written as 
\begin{equation}
    \zeta_t = \pi_\theta(a_0|s_0)p^T(s'_1|s_0,\pi_\theta(s_0))\left(\frac{\pi_\theta(a_0|s_0)p^T(s'_1|s_0,\pi_\theta(s_0))}{\pi^*a(a'_0|s_0)p^S(s'_1|s_0,\pi^*(s_0))}\right).
\end{equation}
Note that above expression is a proper definition of KL divergence where the random variable for two probabilities $p_{\pi_\theta}(.)$ and $q_{\pi^*}(.)$ is the trajectory $\tau = (s_0, s_1, s_2,\dots)$. The KL divergence expression also satisfies the absolute continuity condition owing to fact that the state space for source and target are same.

Computing the KL divergence over the entire trajectory, we can derive the expression for total behavioral adaptation intrinsic return as follows
\begin{eqnarray}
&&\hspace{-10mm}\int_{\tau} p_{\pi_\theta}(\tau)\log\left(\frac{p_{\pi_\theta}(\tau)}{q_{\pi^*}(\tau)}\right)d\tau \nonumber \\
&=&
\displaystyle \mathop{\mathbb{E}}_{s_t \sim \tau}\left(\log\left(\frac{\rho(s_0)\pi(a_0|s_0)p^T(s'_1|s_0,\pi_\theta(s_0))\ldots}{\rho(s_0)\pi^*(a'_0|s_0)p^S(s'_1|s_0,\pi^*(s_0))\dots}\right)\right).\nonumber \\
\label{eq:dkl_logterm}
\end{eqnarray}
\section{Total return gradient with respect to policy parameters}
\label{Gradient derivation}
The total return which we aim to maximize in adapting the source policy to target is the mixture of environmental rewards and Intrinsic KL divergence reward as follows,
\begin{equation}
   \bar{\eta}_{_{KL,\beta}}(\pi_\theta, \pi^*) = \displaystyle \mathop{\mathbb{E}}_{s_t, a_t \sim \tau}\left(p_{\pi_\theta}(\tau)\sum_{t=0}^\mathcal{H} r'_t \right),
   \label{eq:adaptive_learining_objective}
\end{equation}
Taking the expectation over policy and transition distribution we can write the above expression
\begin{equation}
   \bar{\eta}_{_{KL,\beta}}(\pi_\theta, \pi^*) = \displaystyle \mathop{\mathbb{E}}_{s_t \sim p^T, a_t \sim \pi_\theta}\left(\sum_{t=0}^\infty \gamma^t r'_t \right) = V^{\pi_\theta}(s).
   \label{eq:adaptive_learining_objective}
\end{equation}
Using the definition of the state-value function, the above objective function can be re-written as
\begin{equation}
    \bar{\eta}_{_{KL,\beta}}(\pi_\theta, \pi^*) = \sum_a\left(\pi_\theta(a|s)Q^{\pi_\theta}(s,a)\right).
\end{equation}
The adaptive policy update methods work by computing an estimator of the  gradient of the return and plugging it into a stochastic gradient ascent algorithm  
\begin{equation}
    \pi^{*T}_\theta = arg\max_{\pi_\theta \in \Pi}P_{Z^n}(\bar{\eta}_{_{KL,\beta}}).
\end{equation}
$$
\theta = \theta + \alpha \hat{g},
$$
where $\alpha$ is the learning rate and $\hat{g}$ is the empirical estimate of the gradient of the total discounted return $\eta_{_{KL}}$. 

Taking the derivative of the total return term 
\begin{eqnarray}
\nabla_\theta(\bar{\eta}_{_{KL,\beta}}) &=& \nabla_\theta V^{\pi_\theta}(s) = \nabla_\theta\left(\sum_a\left(\pi_\theta(a|s)Q^{\pi_\theta}(s,a)\right)\right), \nonumber \\
\nabla_\theta V^{\pi_\theta}(s)&=&\sum_a \nabla_\theta\pi_\theta(a|s)Q^{\pi_\theta}(s,a) + \sum_a \pi_\theta(a|s)\nabla_\theta Q^{\pi_\theta}(s,a). \nonumber \\
\end{eqnarray}
Using the following definition in above expression,
$$Q^{\pi_\theta}(s_{i},a) = p^T(s_{i+1},|s_i,a)(r + \gamma V^\pi_\theta(s_{i+1})).$$
We can rewrite the gradient to total return over policy $\pi_\theta$ as,
\begin{eqnarray}
&&\hspace{-10mm}\nabla_\theta V^{\pi_\theta}(s_0) \nonumber \\
&=&\sum_a \nabla_\theta\pi_\theta(a|s_0)Q^{\pi_\theta}(s_0,a) \nonumber \\
&+&\sum_{s_1} p^T(s_{1},|s_0,a)\sum_a \pi_\theta(a|s_0)\nabla_\theta (r_0 + \gamma V^\pi_\theta(s_1)). \nonumber \\
\end{eqnarray}
As the reward $r_t$ is independent of $\theta$, we can simplify the above expression and can be re-written as
\begin{eqnarray}
\nabla_\theta V^{\pi_\theta}(s_0)&=&\sum_a \nabla_\theta\pi_\theta(a|s_0)Q^{\pi_\theta}(s_0,a) \nonumber \\
&+&\sum_{s_1} \gamma p^T(s_{1},|s_0,a)\sum_a \pi_\theta(a|s_0)\nabla_\theta V^\pi_\theta(s_1). \nonumber \\
\end{eqnarray}
As we can see the above expression has a recursive property involving term $\nabla_\theta V^{\pi_\theta}(s)$. Using the following definition of a discounted state visitation distribution $d^{\pi_\theta}$ 
\begin{eqnarray}
d^{\pi_\theta}(s_0) &=& \rho(s_0) + \gamma \sum_a \pi(a|s_0)\sum_{s_1}p^T(s_1|s_0,a) \nonumber \\ &&+ \gamma^2 \sum_a \pi(a|s_1)\sum_{s_2}p^T(s_2|s_1,a) \dots
\end{eqnarray}
we can write the gradient of transfer objective as follows,
\begin{equation}
    \nabla_\theta(\eta_{_{KL,\beta}}) = \sum_{s \in \mathcal{S}}d^{\pi_\theta}(s)\sum_{a \in \mathcal{A}}\nabla_\theta\pi_\theta(a|s)Q^{\pi_\theta}(s,a)
    \label{eq:objective_gradient}
\end{equation}
Considering an off-policy RL update, where $\pi_{\theta^-}$ is used for collecting trajectories over which the state-value function is estimated, we can rewrite the above gradient for offline update as follows,

Multiplying and dividing Eq-\ref{eq:objective_gradient} by $\pi_{\theta^-}(a|s)$ and $\pi_{\theta}(a|s)$ we form a gradient estimate for offline update,
\begin{equation}
  = \sum_{s \in \mathcal{S}}d^{\pi_{\theta^-}}(s)\sum_{a \in \mathcal{A}}\pi_{\theta^-}(a|s)\frac{\pi_\theta(a|s)}{\pi_{\theta^-}(a|s)}\frac{\nabla_\theta\pi_\theta(a|s)}{\pi_{\theta}(a|s)} Q^{\pi_{\theta^-}}(s,a)
\end{equation}
where the ratio $\left(\frac{\pi_\theta(a|s)}{\pi_{\theta^-}(a|s)}\right)$ is importance sampling term, and using the following identity the above expression can be rewritten as
\begin{eqnarray}
&&\frac{\nabla_\theta\pi_\theta(a|s)}{\pi_{\theta}(a|s)} = \nabla_\theta \log\pi_\theta(a|s)  \nonumber \\
&\Rightarrow&\displaystyle \mathop{\mathbb{E}}_{s_t \sim d^{\pi_{\theta^-}}, a_t \sim \pi_{\theta^-}}\left(\frac{\pi_\theta(a|s)}{\pi_{\theta^-}(a|s)}Q^{\pi_{\theta^-}}(s,a)\nabla_\theta \log\pi_\theta(a|s)\right) \nonumber\\
\end{eqnarray}   

\subsection{Theoretical bounds on sample complexity }
Although there is some empirical evidence that transfer can improve performance in subsequent reinforcement-learning tasks, there are not many theoretical guarantees in the literature. Since many of the existing transfer algorithms approach the problem of transfer as a method of providing good initialization to target task RL, we can expect the sample complexity of those algorithms to still be a function of the cardinality of state-action pairs $|N| = |\mathcal{S}| \times |\mathcal{A}|$. On the other hand, in a supervised learning setting, the theoretical guarantees of the most algorithms have no dependency on size (or dimensionality) of the input domain (which is analogous to $|N|$ in RL). Having formulated a policy transfer algorithm using labeled reference trajectories derived from optimal source policy, we construct supervised learning like PAC property of the proposed method. For deriving, the lower bound on the sample complexity of the proposed transfer problem, we consider only the adaptation part of the learning i.e., the case when $\beta = 1$. This is because, in ATL, adaptive learning is akin to supervised learning, since the source reference trajectories provide the target states 
given every $(s_t, a_t)$ pair.

Suppose we are given the learning problem specified with training set $Z^n = (Z1, \ldots Z_n)$ where each $Z_i = (\{s_i, a_i\})_{i=0}^n$ are  independently drawn trajectories according to some distribution $P$. Given the data $Z^n$ we can compute the empirical return $P_{Z^n}(\bar{\eta}_{_{KL,\beta}})$ for every $\pi_\theta \in \Pi$,  we will show that the following holds:
\begin{equation}
    \|P_{Z^n}(\bar{\eta}_{_{KL,\beta}}) - P(\bar{\eta}_{_{KL,\beta}})\| \leq \epsilon.
\end{equation}
with probability at least $1-\delta$, for some very small $\delta$ s.t $0 \leq \delta \leq 1$.  We can claim that the empirical return for all $\pi_\theta$ is a sufficiently accurate estimate of the true return function. Thus a reasonable learning strategy is to  find a $\pi_\theta \in \Pi$ that would minimize empirical estimate of the objective 
\begin{eqnarray}
    \pi^{*T}_\theta &=& \arg\max_{\pi_\theta \in \Pi, \beta}\left(\bar{\eta}_{_{KL, \beta}}\right), 
\end{eqnarray}
\begin{theorem}
If the induced class $\mathcal{L}_\Pi$ has uniform convergence in empirical mean property then empirical risk minimization is PAC.
\end{theorem}
For notation simplicity we drop the superscript $T$ (for Target domain) and subscript $\theta$ (policy parameters) in further analysis. Unless stated we will using following simplifications $\hat{\pi}^* = \hat{\pi}^{*T}_\theta$ and ${\pi}^* = {\pi}^{*T}_\theta$
\begin{proof}
Fix $\epsilon, \delta >0$ we will show that for sufficiently large $n \geq n(\epsilon,\delta)$
\begin{equation}
    P^n( P(\bar{\eta}_{_{KL, \hat{\pi}^{*}}}) - P(\bar{\eta}_{_{KL, {\pi^*}}}) \geq \epsilon) \leq \delta
    \label{eq:12}
\end{equation}
Let $\pi^{*} \in \Pi$ be the minimizer of true return $P(\bar{\eta}_{_{KL}})$, further adding and subtracting the terms $P_{Z^n}(\bar{\eta}_{_{KL,\hat{\pi}^{*}}})$ and $P_{Z^n}(\bar{\eta}_{_{KL,{\pi^*}}})$ we can write
\begin{eqnarray}
     &&\hspace{-20mm}P(\bar{\eta}_{_{KL,\hat{\pi}^{*}}})-P(\bar{\eta}_{_{KL, {\pi^*}}})= \nonumber \\
     &&P(\bar{\eta}_{_{KL,\hat{\pi}^{*}}})-P_{Z^n}(\bar{\eta}_{_{KL,\hat{\pi}^{*}}}) \nonumber \\
     &&+P_{Z^n}(\bar{\eta}_{_{KL,\hat{\pi}^{*}}})-P_{Z^n}(\bar{\eta}_{_{KL,{\pi^*}}})\nonumber \\
     &&+P_{Z^n}(\bar{\eta}_{_{KL,{\pi^*}}})-P(\bar{\eta}_{_{KL, {\pi^*}}}) \nonumber \\
\end{eqnarray}
To simplify, the three terms in the above expression can be handled individually as follows,
\begin{enumerate}
    \item $P(\bar{\eta}_{_{KL,\hat{\pi}^{*}}})-P_{Z^n}(\bar{\eta}_{_{KL,\hat{\pi}^{*}}})$
    \item $P_{Z^n}(\bar{\eta}_{_{KL,\hat{\pi}^{*}}})-P_{Z^n}(\bar{\eta}_{_{KL,{\pi^*}}})$
    \item $P_{Z^n}(\bar{\eta}_{_{KL,{\pi^*}}})-P(\bar{\eta}_{_{KL, {\pi^*}}})$
\end{enumerate}
Lets consider the term $P_{Z^n}(\bar{\eta}_{_{KL,\hat{\pi}^*}})-P_{Z^n}(\bar{\eta}_{_{KL,{\pi^*}}})$  in the above expression is always negative semi-definite, since $\hat{\pi}^*$ is a maximizer wrto $P_{Z^n}(\bar{\eta}_{_{KL}})$, hence $P_{Z^n}(\bar{\eta}_{_{KL,\hat{\pi}^*}}) \leq P_{Z^n}(\bar{\eta}_{_{KL,{\pi^*}}})$ always, i.e
$$
P_{Z^n}(\bar{\eta}_{_{KL,\hat{\pi}^*}})-P_{Z^n}(\bar{\eta}_{_{KL,{\pi^*}}}) \leq 0
$$
Next the 1st term can be bounded as 
$$
P(\bar{\eta}_{_{KL,\hat{\pi}^{*}}})-P_{Z^n}(\bar{\eta}_{_{KL,\hat{\pi}^{*}}}) 
\leq \sup_{\pi \in \Pi}[P_{Z^n}(\eta_{_{KL}}) - P(\bar{\eta}_{_{KL}})]$$
$$
\leq \sup_{\pi \in \Pi}\|P_{Z^n}(\bar{\eta}_{_{KL}}) - P(\bar{\eta}_{_{KL}})\|
$$
Similarly upper bound can be written for the 3rd term
Therefore we can upper bound the above expression as
\begin{eqnarray}
P(\bar{\eta}_{_{KL,\hat{\pi}^*}})-P(\bar{\eta}_{_{KL, {\pi^*}}}) \leq
2\sup_{\pi \in \Pi}\|P_{Z^n}(\bar{\eta}_{_{KL}}) - P(\bar{\eta}_{_{KL}})\| \nonumber
\end{eqnarray}
From Equation-(\ref{eq:12}) we have
\begin{equation}
    \sup_{\pi \in \Pi}\|P_{Z^n}(\bar{\eta}_{_{KL}}) - P(\bar{\eta}_{_{KL}})\| \geq \epsilon/2
\end{equation}
Using McDiarmids inequality and union bound, we can state the probability of this event as
\begin{equation}
    P^n(\|P_{Z^n}(\bar{\eta}_{_{KL}}) - P(\bar{\eta}_{_{KL}})\| \geq \epsilon/2) \leq 2|\Pi|e^{-\frac{n\epsilon^2}{2C^2}}
\end{equation}
The finite difference bound
$$
C = \frac{1}{1-\gamma}
$$
Equating the RHS of the expression to $\delta$ and solving for $n$ we get
\begin{equation}
    n(\epsilon,\delta) \geq \frac{2}{\epsilon^2(1-\gamma)^2}\log\left(\frac{2|\Pi|}{\delta}\right)
\end{equation}
for $n \geq n(\epsilon,\delta)$ the probability of receiving a bad sample is less than $\delta$. 
\end{proof}
\subsection{$\epsilon$-Optimality result under Adaptive Transfer-Learning}
Consider MDP $M^*$ and $\hat{M}$ which differ in their transition models. For the sake of analysis,
let $M^*$ be the MDP with ideal transition model, such that target follows source transition $p^*$ precisely. Let $\hat{p}$ be the transition model achieved using the estimated policy learned over data interacting with the target model and the associated MDP be denoted as $\hat{M}$.  We analyze the $\epsilon$-optimality of return under adapted source optimal policy through ATL.

\begin{definition}
\label{defn-1}
Given the value function $V^* = V^{\pi^*}$ and model $M_1$ and $M_2$, which only differ in the corresponding transition models $p_1$ and $p_2$. Define $\forall s,a \in \mathcal{S} \times \mathcal{A}$
$$d_{M_1,M_2}^{V^*} = \sup_{s,a \in \mathcal{S} \times \mathcal{A}}\left\vert \displaystyle \mathop{\mathbb{E}}_{s' \sim P_1(s,a)} [V^{*}(s')] - \displaystyle \mathop{\mathbb{E}}_{s' \sim {P}_2(s,a)} [V^{*}(s')]\right\vert.$$
\end{definition}
\begin{lemma}
\label{lemma:epsilon-optimality}
Given $M^*$, $\hat M$ and value function $V^{\pi^*}_{M^*}$, $V^{\pi^*}_{\hat M}$ the following bound holds 
$\left\Vert V^{\pi^*}_{M^*} -  V^{\pi^*}_{\hat M}\right\Vert_{\infty} \leq \frac{\gamma \epsilon}{(1-\gamma)^2}$
\end{lemma}
where $\max_{s,a}\|\hat{p}(.|s,a)-p^*(.|s,a)\| \leq \epsilon$ and $\hat{p}$ and $p^*$ are transition of MDP $\hat{M}, M^*$ respectively.

The proof of this lemma is based on the simulation lemma \cite{kearns2002near} (see Supplementary document). 
Similar results for RL with imperfect models were reported by \cite{NanJiang}. 
\begin{lemma}
\label{lemma:epsilon-optimality}
Given $M^*$, $\hat M$ and value function $V^{\pi^*}_{M^*}$, $V^{\pi^*}_{\hat M}$ the following bound holds 
$\left\Vert V^{\pi^*}_{M^*} -  V^{\pi^*}_{\hat M}\right\Vert_{\infty} \leq \frac{\gamma \epsilon}{(1-\gamma)^2}$
\end{lemma}
where $\max_{s,a}\|\hat{p}(.|s,a)-p^*(.|s,a)\| \leq \epsilon$ and $\hat{p}$ and $p^*$ are transition of MDP $\hat{M}, M^*$ respectively.

\begin{proof}
For any $s \in \mathcal{S}$
\begin{eqnarray}
&&|V^{\pi^*}_{\hat{M}}(s)-V^{\pi^*}_{M^*}(s)|_\infty \nonumber \\
&&= |r(s,a) + \gamma \left<\hat{p}(s'|s,a), V^{\pi^*}_{\hat{M}}(s')\right> \nonumber \\
&&- r(s,a)-\gamma\left<p^*(s'|s,a), V^{\pi^*}_{M^*}(s')\right>|_\infty\nonumber
\end{eqnarray}

Add and subtract the term $\gamma\left<p^*(s'|s,a), V^{\pi^*}_{\hat{M}}(s')\right>$
\begin{eqnarray}
&=& |\gamma \left<\hat{p}(s'|s,a), V^{\pi^*}_{\hat{M}}(s')\right>-\gamma\left<p^*(s'|s,a), V^{\pi^*}_{\hat{M}}(s')\right> \nonumber \\
&+&\hspace{-2mm}\gamma\left<p^*(s'|s,a), V^{\pi^*}_{\hat{M}}(s')\right> -\gamma\left<p^*(s'|s,a), V^{\pi^*}_{M^*}(s')\right>|_\infty\nonumber\\
&\leq&\hspace{-2mm}\gamma |\left<\hat{p}(s'|s,a), V^{\pi^*}_{\hat{M}}(s')\right>-\left<p^*(s'|s,a), V^{\pi^*}_{\hat{M}}(s')\right>| \nonumber \\
&+&\hspace{-2mm}\gamma|\left<p^*(s'|s,a), V^{\pi^*}_{\hat{M}}(s')\right> -\gamma\left<p^*(s'|s,a), V^{\pi^*}_{M^*}(s')\right>|_\infty\nonumber\\
&\leq&\hspace{-2mm}\gamma|\hat{p}(s'|s,a) - p^*(s'|s,a)|_\infty|V^{\pi^*}_{\hat{M}}(s')|_\infty \nonumber \\
&&+ \gamma|V^{\pi^*}_{\hat{M}}(s)-V^{\pi^*}_{M^*}(s)|_\infty \nonumber
\end{eqnarray}

Using the definition of $\epsilon$ in above expression, we can write
$$
|V^{\pi^*}_{\hat{M}}(s)-V^{\pi^*}_{M^*}(s)|_\infty \leq \gamma\epsilon|V^{\pi^*}_{\hat{M}}(s')|_\infty + \gamma|V^{\pi^*}_{\hat{M}}(s)-V^{\pi^*}_{M^*}(s)|_\infty
$$
Therefore
$$
|V^{\pi^*}_{\hat{M}}(s)-V^{\pi^*}_{M^*}(s)|_\infty \leq \frac{\gamma\epsilon|V^{\pi^*}_{\hat{M}}(s')|_\infty}{1-\gamma}
$$
Now we solve for expression $|V^{\pi^*}_{\hat{M}}(s')|_\infty$. We know that this term is bounded as
$$
|V^{\pi^*}_{\hat{M}}(s')|_\infty \leq \frac{R_{max}}{1-\gamma}
$$
where $R_{max} = 1$, therefore we can write the complete expression as
$$
|V^{\pi^*}_{\hat{M}}(s)-V^{\pi^*}_{M^*}(s)|_\infty \leq \frac{\gamma\epsilon}{(1-\gamma)^2}
$$
\end{proof}
\begin{table}[tbh!]
    \centering
    \begin{tabular}{|c|c|c|c|c|}
    \hline
     Env&Property&source&Target&\%Change\\
     \hline
     Hopper&Floor Friction&1.0&2.0&+100\%\\
     \hline
     HalfCheetah&gravity&-9.81&-15&+52\%\\
     &Total Mass&14&35&+150\%\\
     &Back-Foot Damping&3.0&1.5&-100\%\\
     &Floor Friction&0.4&0.1&-75\%\\
     \hline
     Walker2d&Density&1000&1500&+50\%\\
     &Right-Foot Friction&0.9&0.45&-50\%\\
     &Left-Foot Friction&1.9&1.0&-47.37\%\\
     \hline
\end{tabular}
    \caption{Transition Model and environment properties for Source and Target task and \% change}
    \label{Table:env_property}
\end{table}
\begin{table}[tbh!]
    \centering
    \begin{tabular}{|c|ccc|}
         \hline
            & Hopper&Walker2d& HalfCheetah \\
         State Space& 12&18&17\\
         Control Space& 3&6&6\\
         Number of layers&3&3&3\\
         Layer Activations&tanh&tanh&tanh\\
         Total num. of network params&10530&28320 &26250\\
         \hline
         Discount&0.995&0.995&0.995\\
         Learning rate ($\alpha$)&$1.5\times10^{-5}$&$8.7\times10^{-6}$&$9\times10^{-6}$\\
         $\beta$ initial Value&0.5&0.5&0.5\\
         $\beta$-Learning rate ($\bar{\alpha}$)&0.1&0.1&0.1\\
         Batch size&20&20&5\\
         Policy Iter&3000&5000&1500\\
         \hline
    \end{tabular}
    \caption{Policy Network details and Network learning parameter details}
    \label{tab:network_and Learning_details}
\end{table}
\section{Learning the Mixing Coefficient $\beta$}
A hierarchical update of the mixing coefficient $\beta$ is carried out over  n-test trajectories, collected using the updated policy network $\pi_{\theta'}(a|s)$. 
The mixing coefficient $\beta$ is learnt by optimizing the return over trajectory as follows,
$$
\beta = arg\max_\beta(\bar{\eta}_{KL,\beta}(\pi_{\theta'}, \pi^*))
$$
where $\theta'$ is parameter after the policy update step. 
$$
\beta = arg\max_\beta\mathop{\mathbb{E}}_{s_t,a_t \sim \tau}\left(p_{\pi_\theta}(\tau)\sum_{t=1}^\infty \gamma^tr'_t\right)
$$
We can use gradient ascent to update parameter $\beta$ in direction of optimizing the reward mixing as follows,
$$
\beta \leftarrow \beta+\bar{\alpha}\nabla_\beta(\bar{\eta}_{KL,\beta}(\pi_{\theta'}, \pi^*)).
$$
Using the definition of mixed reward as $r'_t = (1-\beta)r_t - \beta\zeta_t$, we can simplify the above gradient as,
$$
\beta \leftarrow \beta+\bar{\alpha}\mathop{\mathbb{E}}_{s_t,a_t \sim \tau}\left(p_{\pi_\theta}(\tau)\sum_{t=1}^\infty \gamma^t\nabla_\beta(r'_t)\right)
$$
$$
\beta \leftarrow \beta+\bar{\alpha}\mathop{\mathbb{E}}_{s_t,a_t \sim \tau}\left(\sum_{t=1}^\infty \gamma^t(r_t - \zeta_t)\right).
$$
We use stochastic gradient ascent to update the mixing coefficient $\beta$ as follows
$$
\beta \leftarrow \beta + \bar{\alpha} \hat{g}_\beta,
\hspace{2mm}
s.t \hspace{2mm} 0 \leq \beta \leq 1 .
$$
where $\bar{\alpha}$ is the learning rate and $\hat{g}_\beta = P_{Z^n_{test}}(\nabla_\beta \bar{\eta}_{_{KL, \beta}})$ is the empirical estimate of the gradient of the total return $\bar{\eta}_{_{KL, \beta}}(\pi_{\theta'}, \pi^*)$. The gradient estimate $\hat{g}_\beta$ over data $(Z^n_{test} :\{s_i,a_i,a'_i\}_i^T)$ is computed as follows,
$$
\hat{g}_\beta = \frac{1}{N}\sum_{i=1}^N\left(\sum_{t=1}^\mathcal{H} \gamma^t(r_t - \zeta_t)\right)
$$
where $\mathcal{H}$ truncated trajectory length from experiments.

As we can see the gradient of objective with respect to mixing coefficient $\beta$ is an average over difference between environmental and intrinsic rewards. If $r_t-\zeta_t \geq 0$ the update will move parameter $\beta$ towards favoring learning through exploration more than learning through adaptation and visa versa.

As $\beta$ update is a constrained optimization with constraint $0 \leq \beta \leq 1$. We handle this constrained optimization by modelling $\beta$ as output of Sigmoidal network parameterized by parameters $\phi$.
$$
\beta = \sigma(\phi)
$$
And the constrained optimization can be equivalently written as optimizing w.r.to $\phi$ as follows
$$
\phi \leftarrow \phi + \bar{\alpha} \hat{g}_\beta \nabla_\phi(\beta),
\hspace{2mm}
where \hspace{2mm} \beta = \sigma(\phi)
$$
The reward mixing co-efficient $\beta$ learned for HalfCheetah, Hopper and Walker2d envs is provided in Figure-\ref{fig:beta_plot}. For all the experiments we start with $\beta = 0.5$ that is placing equal probability of learning through adaptation and learning through exploration. As we can observe the reward mixing leans towards learning through adaptation for HalfCheetah and Hopper envs. Whereas, as for Walker2d the beta initially believes learning from exploration more, but quickly leans toward learning from source policy and adaptation. 
\end{document}